\documentclass[hidelinks]{ieeetj}

\usepackage{amsmath,amssymb,amsfonts}
\usepackage{algorithmic}
\usepackage{graphicx,color}
\usepackage{textcomp}

\usepackage{multirow}%
\usepackage{amsthm}%
\usepackage{mathrsfs}%
\usepackage{xcolor}%
\usepackage{booktabs}%

\usepackage{array}
\usepackage{bm}
\usepackage[colorlinks=true,allcolors=blue]{hyperref}
\usepackage[super]{nth}
\usepackage{orcidlink}
\usepackage{siunitx}
\usepackage{subfig}
\usepackage{tabularx}
\usepackage{tikz}
\usepackage{tikz-3dplot}
\usetikzlibrary{angles,arrows.meta,babel,calc,decorations.markings,decorations.pathmorphing,fit,patterns,positioning,quotes,shapes}
\usepackage{schemabloc}
\usepackage{todonotes}
\usepackage{varwidth}

\usepackage{cleveref}

\newcommand\wordcount{
    \immediate\write18{texcount -merge -sub=section \jobname.tex  | grep "Section" | sed -e 's/+.*//' | sed -n \thesection p > 'count/count_\thesection.txt'}
(\input{count/count_\thesection.txt}words)}

\newcommand\totalwordcount{
    \immediate\write18{texcount -merge \jobname.tex | grep "Words in text:" | sed -e 's/[^0-9]*//g' > count/total_count.txt}
    (Total words: \input{count/total_count.txt} words)}

\renewcommand{\vec}[1]{\bm{#1}}

\newcommand{\rev}[1]{{\color{black} #1}}

\crefname{equation}{}{}
\Crefname{equation}{}{}
\crefname{figure}{Figure}{Figure}
\Crefname{figure}{Figure}{Figure}
\crefname{table}{Table}{Table}
\Crefname{table}{Table}{Table}
\crefname{section}{Section}{Section}
\Crefname{section}{Section}{Section}

\def\BibTeX{{\rm B\kern-.05em{\sc i\kern-.025em b}\kern-.08em
    T\kern-.1667em\lower.7ex\hbox{E}\kern-.125emX}}
\AtBeginDocument{\definecolor{tmlcncolor}{cmyk}{0.93,0.59,0.15,0.02}\definecolor{NavyBlue}{RGB}{0,86,125}}

\usepackage[sorting=none]{biblatex}
\def\OJlogo{\vspace{-4pt}$<$Society logo(s) and publication title will appear here.$>$}
\def\seclogo{\vspace{10pt}$<$Society logo(s) and publication title will appear here.$>$}

\def\authorrefmark#1{\ensuremath{^{\textbf{#1}}}}

\begin{document}
\receiveddate{XX Month, XXXX}
\reviseddate{XX Month, XXXX}
\accepteddate{XX Month, XXXX}
\publisheddate{XX Month, XXXX}
\currentdate{XX Month, XXXX}
\doiinfo{\href{https://doi.org/10.1109/TFR.2025.3632773}{TFR.2025.3632773}}

\markboth{Botany Meets Robotics in Alpine Scree Monitoring}{Davide De Benedittis {et al.}}

\title{Botany Meets Robotics in \\Alpine Scree Monitoring}

\author{
Davide~De~Benedittis${}^{\orcidlink{0000-0002-7185-6784}}$\authorrefmark{1,2} (Student Member, IEEE),
Giovanni~Di~Lorenzo${}^{\orcidlink{0009-0009-6640-1927}}$\authorrefmark{1,2},
Franco~Angelini${}^{\orcidlink{0000-0003-2559-9569}}$\authorrefmark{1,2} (Member, IEEE),
Barbara~Valle${}^{\orcidlink{0000-0003-4829-4776}}$\authorrefmark{3},
Marina~Serena Borgatti${}^{\orcidlink{0000-0003-3842-8646}}$\authorrefmark{4},
Paolo~Remagnino${}^{\orcidlink{0000-0002-9168-7746}}$\authorrefmark{5} (Member, IEEE),
Marco~Caccianiga${}^{\orcidlink{0000-0001-9715-1830}}$\authorrefmark{4},
Manolo~Garabini${}^{\orcidlink{0000-0002-5873-3173}}$\authorrefmark{1,2}~(Member, IEEE)
}

\affil{Department of Information Engineering, University of Pisa, Pisa, Italy}
\affil{Research Center E. Piaggio, School of Engineering, University of Pisa, Pisa, Italy}
\affil{Department of Life Sciences, University of Siena, Siena, Italy}
\affil{Department of Life Sciences, University of Milan, Milano, Italy}
\affil{Department of Computer Science, Durham University, Durham, United Kingdom}
\corresp{Corresponding author: Davide De Benedittis (e-mail: davide.debenedittis@phd.unipi.it).}
\authornote{This work was supported in part by the European Union's Horizon 2020 Research and Innovation Programme under Grant Agreement No. 101016970 (Natural Intelligence); in part by the European Union's Horizon MSCA-2023-SE-01-01 Staff Exchanges 2023 Program under Grant Agreement No. 101182891 (NEUTRAWEED); and in part by the Italian Ministry of Education and Research in the framework of the
“FoReLab” (Future-oriented Research Lab) Project (Departments of Excellence), and in part by the European Union Next Generation EU project ECS00000017 "Ecosistema dell'Innovazione" Tuscany Health Ecosystem (THE, PNRR, Spoke 9: Robotics and Automation for Health).}

\begin{abstract}
According to the European Union's Habitat Directive, habitat monitoring plays a critical role in response to the escalating problems posed by biodiversity loss and environmental degradation.
Scree habitats, hosting unique and often endangered species, face severe threats from climate change due to their high-altitude nature.
Traditionally, their monitoring has required highly skilled scientists to conduct extensive fieldwork in remote, potentially hazardous locations, making the process resource-intensive and time-consuming.
This paper presents a novel approach for scree habitat monitoring using a legged robot to assist botanists in data collection and species identification.
Specifically, we deployed the ANYmal C robot in the Italian Alpine bio-region in two field campaigns spanning two years and leveraged deep learning to detect and classify key plant species of interest.
Our results demonstrate that agile legged robots can navigate challenging terrains and increase the frequency and efficiency of scree monitoring.
When paired with traditional phytosociological surveys performed by botanists, this robotics-assisted protocol not only streamlines field operations but also enhances data acquisition, storage, and usage.
The outcomes of this research contribute to the evolving landscape of robotics in environmental science, paving the way for a more comprehensive and sustainable approach to habitat monitoring and preservation.
\end{abstract}


\IEEEoverridecommandlockouts
\IEEEpubid{\makebox[\columnwidth]{Electronic ISSN: 2997-1101~\copyright2025 IEEE \hfill}
\hspace{\columnsep}\makebox[\columnwidth]{ }}

\maketitle

\IEEEpubidadjcol

\section{Introduction}\label{sec:introduction}

\begin{figure*}[!ht]
    \centering
    \includegraphics[width=.8\textwidth]{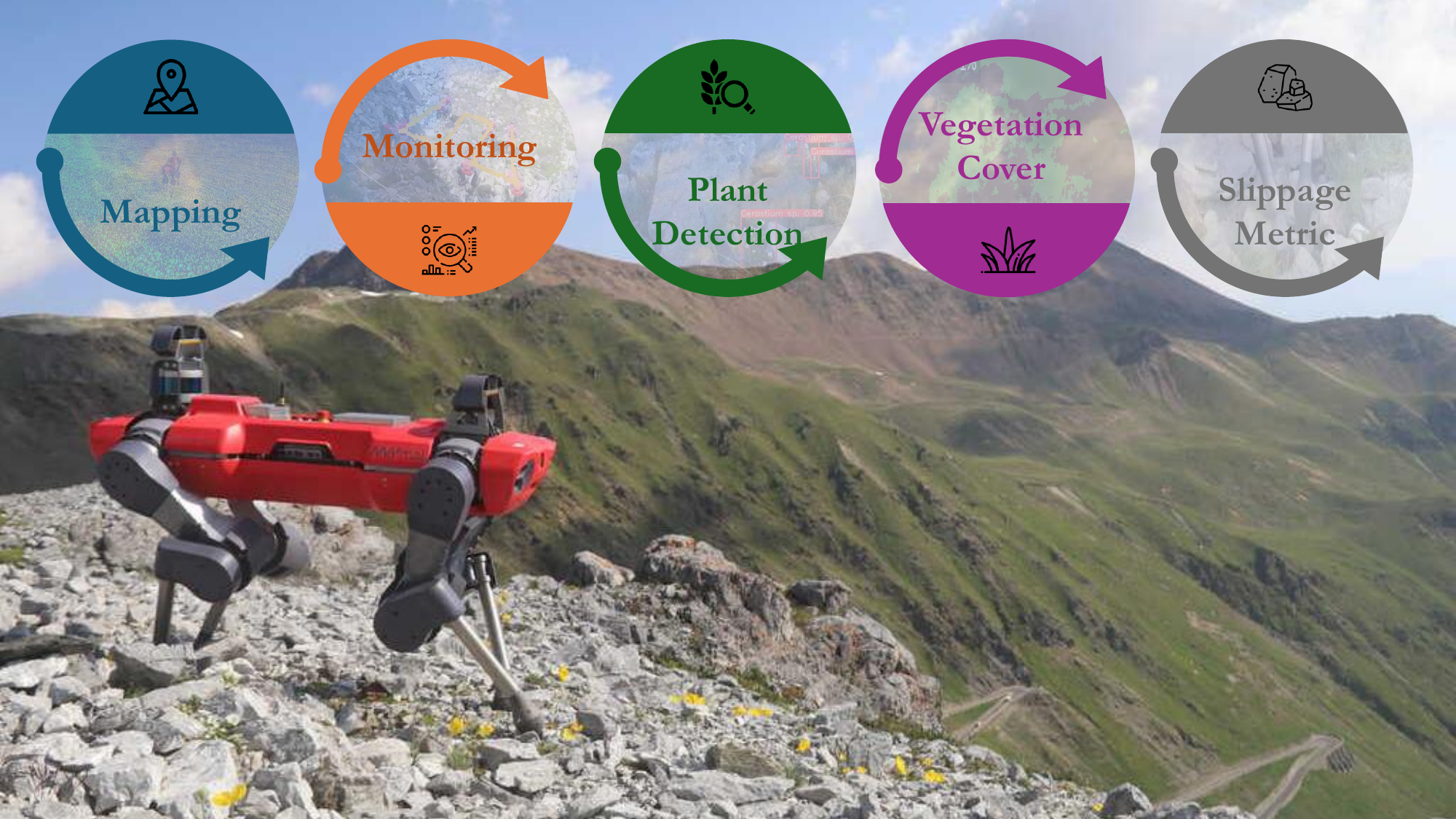}
    \caption{Overview of the proposed monitoring framework.}
    \label{fig:cover_pic}
\end{figure*}











\tikzstyle{block} = [draw, rectangle, minimum height=3em, rounded corners, minimum width=6em]

\begin{figure*}
    \centering
    \begin{tikzpicture}[auto, node distance=2cm, >=latex']
    \usetikzlibrary{shapes.misc} 

    \node[block, label={\textbf{Mapping}}, minimum width=5em, inner sep=0pt] (mapping)
    {\begin{tikzpicture}
        \node[rounded corners, clip, inner sep=0pt] {\includegraphics[height=6em]{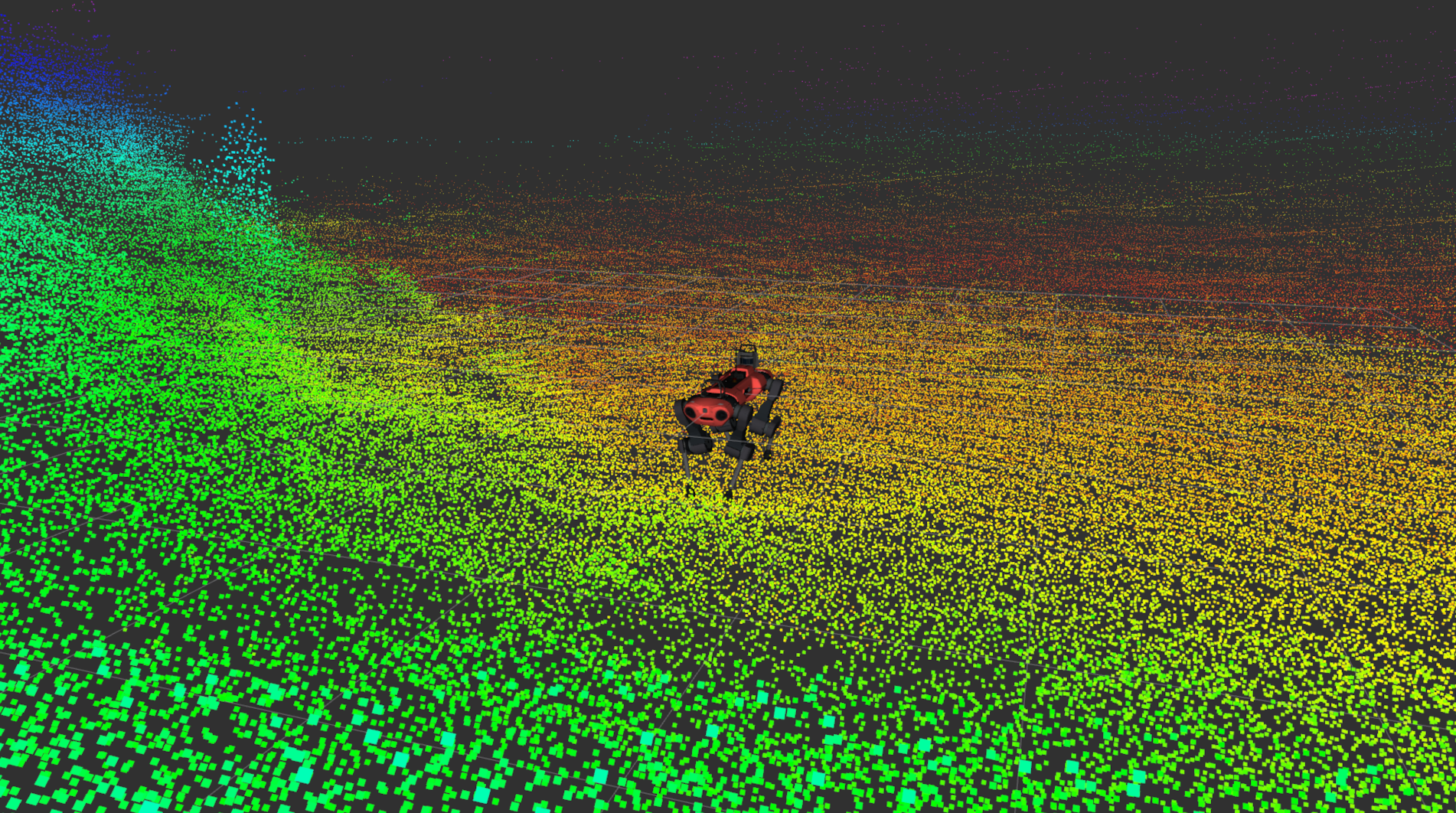}};
    \end{tikzpicture}};
    
    \node[block, label={\textbf{Mission Plan}}, inner sep=0pt, right=3.0cm of mapping] (mission_plan)
    {\begin{tikzpicture}
        \node[rounded corners, clip, inner sep=0pt] {\includegraphics[height=6em]{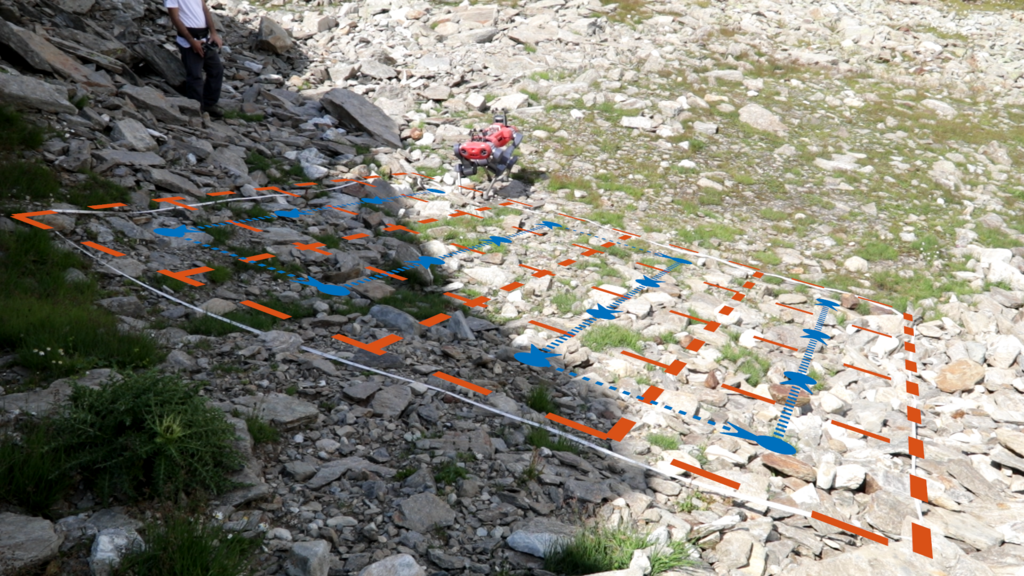}};
    \end{tikzpicture}};

    
    \node[block, minimum size=1cm, below=1cm of mapping] (prediction) {%
        \begin{tabular}{c}
            \small
            \textbf{Prediction} \\
            \tiny
            $\begin{aligned}
            x_{k|k-1} &= F\,x_{k-1|k-1},\\
            P_{k|k-1} &= F\,P_{k-1|k-1}\,F^\top + Q
            \end{aligned}$
            \normalsize
        \end{tabular}
    };
    
    \node[block, below=0.1cm of prediction, minimum size=1cm] (update) {%
        \begin{tabular}{c}
            \small
            \textbf{Update} \\
            \tiny
            $\begin{aligned}
                K_k &= P_{k|k-1}\,H^\top 
                \bigl(H\,P_{k|k-1}\,H^\top + R\bigr)^{-1},\\
                x_{k|k-1} &= x_{k|k-1} + 
                K_k \bigl(z_k - H\,x_{k|k-1}\bigr),\\
                P_{k|k-1} &= \bigl(I - K_k\,H\bigr)\,P_{k|k-1}
            \end{aligned}$
            \normalsize
        \end{tabular}
    };
    
    \draw[->]
        (prediction.east) to[out=340, in=20, looseness=1.5] node[midway, right] (curv_right) {} (update.east);
    \draw[->]
        (update.west) to[out=160, in=200, looseness=1.5] node[midway, left] (curv_left) {} (prediction.west);

    \node[block, label=below:{\textbf{Localization}},fit=(prediction)(update)(curv_left)(curv_right)] (localization) {};
    
    \node[block, below=1.6cm of mission_plan, label=below:{\textbf{Planner}}, inner sep=0pt] (planner) {
        \begin{tikzpicture}
            \node[rounded corners, clip, inner sep=0pt] {\includegraphics[height=6em]{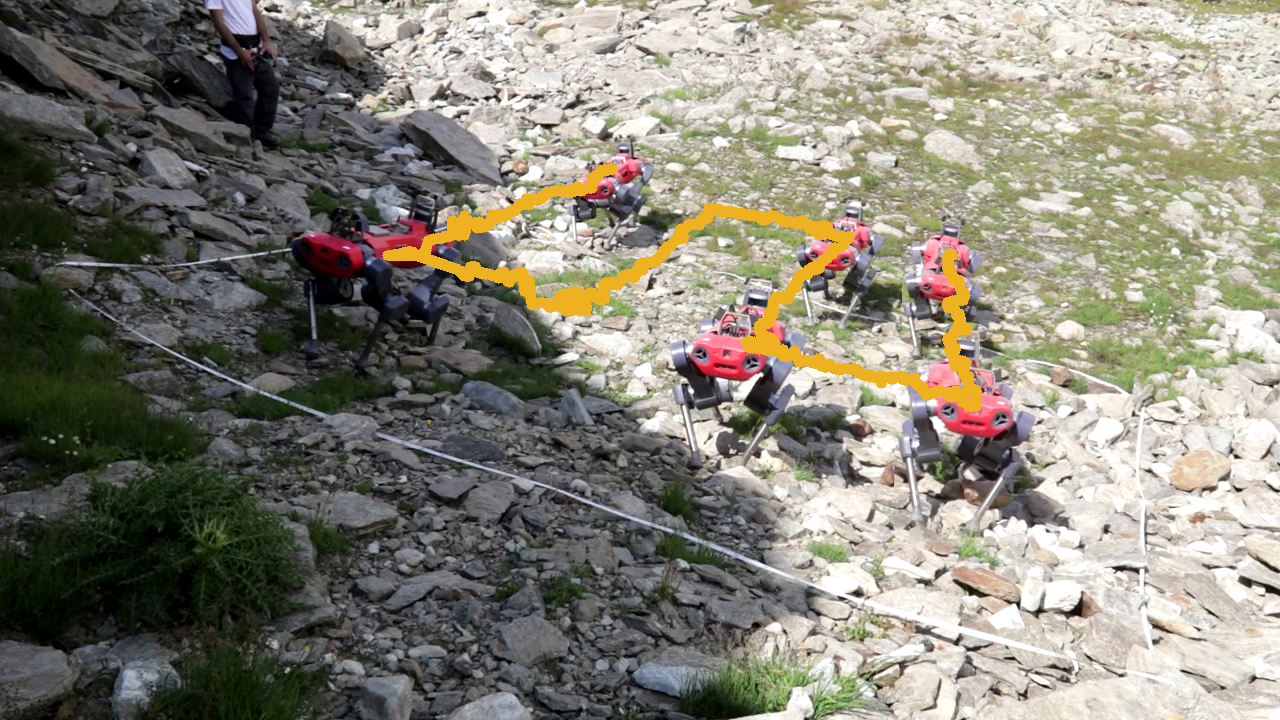}};
        \end{tikzpicture}
    };
    
    \draw [->] (mapping) -- node[midway, right] {map} (localization);
    \draw [->] (mission_plan) -- node[midway, right] {waypoint} (planner);
    
    \node (center_between) at ($(localization)!0.5!(planner)$) {};

    \node[block, label={\textbf{Control}}, below=2.5cm of center_between] (control) {
        \begin{tikzpicture}
            \node[rounded corners, clip, inner sep=0pt] {\includegraphics[height=6em]{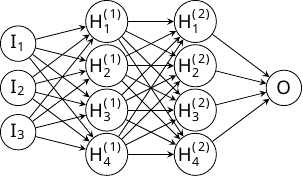}};
        \end{tikzpicture}
    };
    
    \draw [->] (localization) -- node[midway, right] {$\overline{q}, \overline{v}$} (control);
    \draw [->] (planner) -- node[midway, right] {$v_\mathrm{ref}$} (control);
    
    \node[block, label=below:{\textbf{System}}, minimum width=4em, inner sep=0pt, below=1cm of control] (system)
    {\begin{tikzpicture}
        \node[rounded corners, clip, inner sep=0pt] {\includegraphics[height=6em]{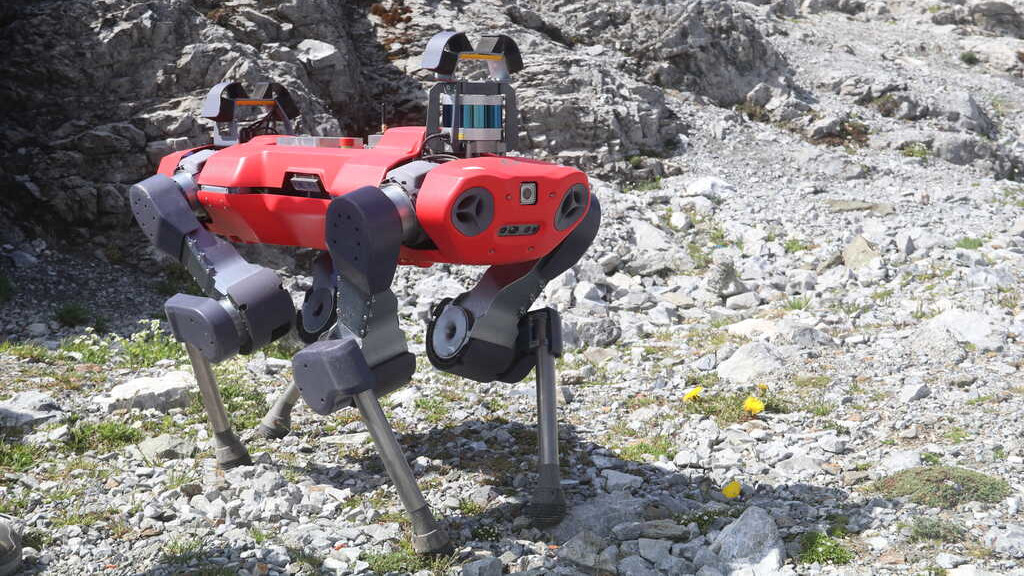}};
    \end{tikzpicture}};
    
    \draw [->] (control) -- node[midway, right] {$q_{\text{ref}}$} (system);

    \draw [->] (system.145) -- (localization.225);

    \node[block, left=2cm of system, label={\textbf{Cameras}}, inner sep=0pt] (camera) {
        \begin{tikzpicture}
            \node[rounded corners, clip, inner sep=0pt] {\includegraphics[height=6em]{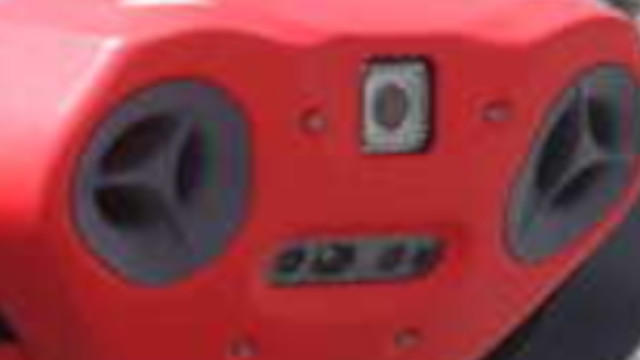}};
        \end{tikzpicture}
    };
    
    \node[block, right=2cm of system, label={\textbf{Odometry}}, inner sep=0pt] (odometry) {
        \begin{tikzpicture}
            \node[rounded corners, clip, inner sep=0pt] {\includegraphics[height=6em]{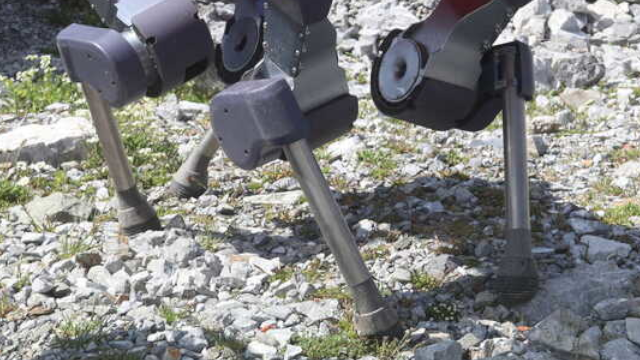}};
        \end{tikzpicture}
    };
    
    \draw [->] (system) -- (camera);
    \draw [->] (system) -- (odometry);
    
    \node[block, below=1cm of camera, inner sep=0, label=below:{\textbf{Plants Detection}}] (plant_det) {
        \begin{tikzpicture}
            \node[rounded corners, clip, inner sep=0pt] {\includegraphics[height=6em]{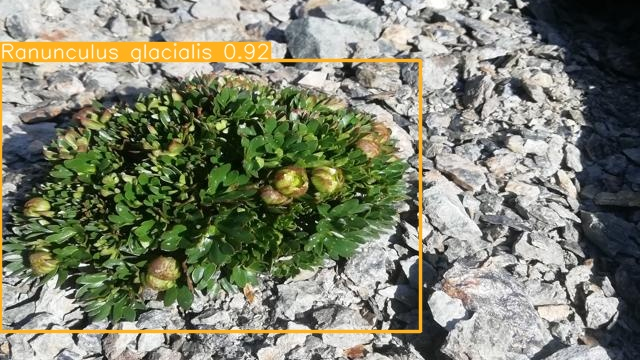}};
        \end{tikzpicture}
    };
    
    \node[block, below=1cm of system, inner sep=0, label=below:{\textbf{Vegetation Cover}}] (veg_cov) {
        \begin{tikzpicture}
            \node[rounded corners, clip, inner sep=0pt] {\includegraphics[height=6em]{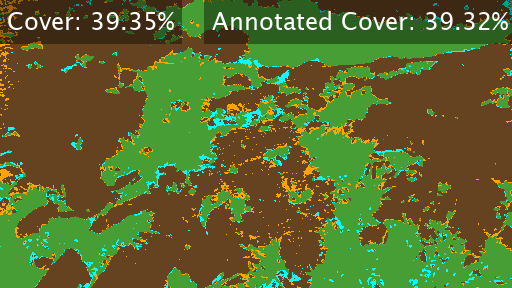}};
        \end{tikzpicture}
    };
    
    \node[block, below=1cm of odometry, inner sep=0, label=below:{\textbf{Slippage Metric}}] (slippage) {
        \begin{tikzpicture}
            \node[rounded corners, clip, inner sep=0pt] {\includegraphics[height=6em]{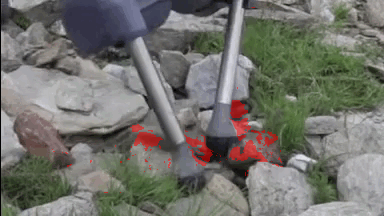}};
        \end{tikzpicture}
    };
    
    \draw [->] (camera) -- (plant_det);
    \draw [->] (camera) -- (veg_cov);
    \draw [->] (odometry) -- node {$q, v, \tau$} (slippage);
    
    \end{tikzpicture}
    \caption{Block diagram representing the proposed monitoring framework.\label{fig:block_diag}}
\end{figure*}

\IEEEPARstart{R}{ecent} years have seen a massive increase in the awareness towards the environmental problems caused by humanity.
Human activities have altered the Earth's climate~\rev{\cite{change2007climate}}, damaged entire habitats~\rev{\cite{watson2019summary}}, and drastically reduced biodiversity~\rev{\cite{ipbes2019global}}.
In response, multiple strategies have been adopted to mitigate these impacts, such as reducing emissions and performing regular environmental monitoring to track changes in ecosystem health.

Natura 2000 is one of the European Union's key instruments for protecting biodiversity, as mandated by the Habitat Directive (European Directive 92/43/EC ``Habitats''~\cite{directive1992council}).
This network of protected areas spreads across all of Europe, encompassing 18\% of land area and 7\% of marine area.
Habitats included in this network are preserved, and their conservation status is continuously monitored.

Among the habitats covered by Natura 2000, scree habitats are particularly noteworthy for hosting endemic and rare species, and their preservation is paramount for avoiding biodiversity loss.
Similarly to other high-altitude habitats, they are particularly threatened by climate change~\cite{nitzu2014scree} and monitoring them offers indirect insights into it~\cite{fragniere2020climate}.

Despite their ecological importance, scree habitats are difficult to monitor.
Current monitoring approaches are both time-consuming and reliant on highly trained experts who must conduct extensive field surveys under hazardous conditions.
Additionally, narrow seasonal windows suitable for monitoring, limited manpower, and challenging terrains have contributed to a dissatisfactory implementation of the Natura 2000 network~\cite{kati2015challenge}, with low spatial coverage and infrequent monitoring.
Moreover, traditional field surveys depend on subjective estimations, such as vegetation cover and debris mobility, introducing variability in data quality and consistency.

Conventional scree monitoring involves two phases:
(1)~a preliminary plot setup, where a rectangular region of at least $\SI{16}{m^2}$ is delimited and its location saved, and
(2)~a detailed survey of the plot performed by the botanist.
Under the Habitat Directive, Typical Species (TS) and Early Warning Species (EWS) serve as indicators of the habitat's conservation status.
The specific survey protocols vary across countries in the EU, with some performing a survey based only on TS and EWS and others performing a more informative but time-intensive phytosociological survey.
\rev{
Phytosociological analysis is a method used in vegetation science to classify and describe plant communities based on their species composition and relative abundance, following standardized protocols such as the Braun-Blanquet approach.
For more details, see \cite{dengler2016phytosociology}.
}

The improvement and automation of habitat monitoring have captured the attention of the research community in recent years~\cite{van2018dawning}.
Although remote sensing offers extensive coverage with very low costs, it cannot achieve the high spatial resolution of in situ monitoring required for monitoring purposes~\cite{lathrop2014comparison}.
Robotic systems, by contrast, have the potential to revolutionize habitat monitoring by enabling in-depth, high-frequency, and cost-effective in situ data collection, thereby complementing traditional methods.

In this work, we develop and propose a framework for scree monitoring using the quadrupedal robot ANYmal C (in~\cref{fig:cover_pic}).
A block diagram of the proposed monitoring framework is shown in~\cref{fig:block_diag}.
The robot assists botanists during data acquisition and autonomously analyzes the collected data, thereby improving the efficiency and data quality acquired during fieldwork.
The main contributions of this paper are:
\rev{
\begin{enumerate}
    \item the development of a monitoring framework for scree habitats based on legged robots (\cref{sec:rob_mon_mission}) that directly addresses the issues of traditional monitoring and other robotic platforms. The framework is composed of:
    \begin{itemize}
        \item a robot-centric quantitative slippage metric (\cref{sec:slippage_metric}) to assess scree mobility,
        \item a vegetation cover estimation method (\cref{sec:veg_cov_est}),
        \item a specialized neural network fine-tuned for scree vegetation which detects the TS and EWS from camera images (\cref{sec:detection_nn}),
    \end{itemize}
    \item a pioneering demonstration of the usage of quadrupeds for environmental monitoring of scree habitats. It validates the proposed monitoring framework in the Italian Alpine bioregion, with experimental results acquired over two years of field campaigns and presented in \cref{sec:results}.
\end{enumerate}
}

The paper is organized as follows.
\Cref{sec:related_work} reviews the state of the art in environmental monitoring, focusing on robots for terrestrial monitoring.
\rev{\Cref{sec:scree_mon} outlines the monitoring process, covering a description of the traditional method (\cref{sec:trad_scree_mon}), the habitats studied in this work (\cref{sec:situ_desc}).
\Cref{sec:rob_mon} first introduces the ANYmal C robot (\cref{sec:rob_equip}) and later expands} on the monitoring mission (\cref{sec:rob_mon_mission}) and \rev{introduces} the slippage metric (\cref{sec:slippage_metric}), the vegetation cover estimation (\cref{sec:veg_cov_est}), and the plant detection neural network (\cref{sec:detection_nn}).
The results are presented in \cref{sec:results}, including locomotion in the scree habitat (\cref{sec:loc_scree_hab}), the benefits of robotic monitoring (\cref{sec:efficiency_rob_mon}), and neural network performance (\cref{sec:vegetation_detection}).
Finally, \cref{sec:discussion} discusses the results, and \cref{sec:conclusion} summarizes the main contributions and outlines future work.
\section{Related Work}\label{sec:related_work}

In this section, we provide an overview of the state of the art of robots for habitat monitoring.
We focus exclusively on robots for terrestrial monitoring because maritime habitats involve completely different problems and solutions.

Unmanned Aerial Vehicles (UAVs) were among the first robots used for environmental monitoring~\cite{dunbabin2012robots} due to their low cost and ease of use.
UAVs have been employed in multiple tasks and habitats, ranging from biodiversity study of flora~\cite{puliti2015inventory} and fauna~\cite{koh2012dawn} to assessing habitats' health status~\cite{michez2016classification} and estimating forests' biomass~\cite{paneque2014small}.
Additionally, they have been used in a wide variety of habitats, from glaciers~\cite{seier2017uas} to savannas~\cite{mayr2018disturbance}, using both passive (e.g., cameras~\cite{woodget2017drones}) and active (e.g., lidar~\cite{sankey2017uav}) sensors.
Nevertheless, UAVs suffer from limited autonomy, reduced capacity for close-up observation in dense vegetation, and vulnerability to harsh weather conditions, which can affect both data quality and safety.
Consequently, UAVs excel at large-scale remote sensing applications\rev{, where they can fully exploit their mobility and speed to cover large areas quickly and efficiently~\cite{bhardwaj2016uavs}.
On the other hand, they are less effective for in-situ close-up monitoring, where they struggle to capture detailed data, disrupt the wildlife~\cite{afridi2025impact}, and can experience difficulty when operating near the ground.}

Wheeled robots are yet another solution that has been tested for monitoring purposes.
The Nomad robot~\cite{bapna1998atacama} and ROBOVOLC~\cite{muscato2003robovolc} are two examples, having been deployed in the Atacama desert in Chile and in volcanic regions, respectively.
Compared to UAVs, wheeled robots offer much longer autonomy, can remain stationary without consuming energy, and can carry high payloads.
However, wheel-based motion is ill-suited for rough and unstructured environments.
Therefore, wheeled robots can be successfully deployed only in less challenging natural environments.

\rev{
Continuous monitoring through fixed stations is yet another possible solution.
However, since habitat monitoring needs to be performed in different locations over time, fixed monitoring stations are not suitable for this task.
}
\rev{Similar to fixed stations, in}~\cite{notomista2019slothbot}, \rev{Notomista et al.} present SlothBot, a cable-based robot designed for long-term forest monitoring.  
It relies on overhead cables installed on tree branches and solar power, enabling extended operation with minimal energy consumption.  
While promising for some forest environments, this method requires a labor-intensive cable installation, which is costly, alters the environment, and is restricted to habitats that can accommodate such infrastructure.
\rev{Additionally, like with fixed stations, it would prove to be very difficult to preserve the stations in a hostile and dynamic environment such as scree habitats.
Therefore, these solutions are not suitable for scree monitoring.}

Legged robots represent a promising alternative to aerial and wheeled robots.
They balance mobility and battery life, achieving better autonomy than most UAVs and better adaptability on complex terrains than wheeled robots.
Dante II~\cite{bares1999dante} was the first legged robot to be developed for monitoring purposes and tested in an Alaskan volcano in 1994.
It paved the way for modern legged robots and demonstrated both their potential and limitations.
However, only recently have the capabilities of legged robots improved enough to allow efficient habitat monitoring to be feasible.
For instance, robots such as HyQ~\cite{semini2011design}, Spot, ANYmal~\cite{hutter2016anymal}, or Unitree quadrupeds have repeatedly proved their agility in unstructured environments.
In~\cite{angelini2023robotic}, \rev{Angelini et al. present a perspective work on} the ``Natural Intelligence approach'' for robotic habitat monitoring.
This is obtained through the combination of the environment, the robot's body, and the mind, i.e., planning, control, and data analysis.
\rev{This work paves the way for the usage of legged robots in habitat monitoring, highlighting their potential and the need for further research in this area.}

The research community has pushed the development of legged robots toward locomotion on highly unstructured terrains, including mountains, sand, forests, and snow.
These efforts have resulted in the development of both hardware and software solutions that allow legged robots to move reliably in challenging environments.
For instance, in \cite{ranjan2023design}, \rev{Ranjan et al.} propose a novel hoof-inspired foot for reducing slippage on challenging terrains.
In~\cite{zhang2023optimal}, \rev{Zhang et al.} present an optimal control strategy specifically designed for wheeled-legged robots traversing mountainous terrains.
Conversely, \cite{debenedittis2025soft} and \cite{choi2023learning} propose a model-based and a reinforcement-learning-based controller for locomoting on compliant terrains, respectively.
\rev{In \cite{ding2023robust}, Ding et al. explore alternative locomotion gaits using model-based articulated soft-robots control~\cite{chhatoi2023optimal}.
Margolis et al. also investigate different gait patterns and characteristics, and achieve more performant locomotion by manually changing these parameters~\cite{margolis2023walk}.
Finally, \cite{miki2022learning,cheng2024quadruped} present frameworks for locomotion using perception and for dealing with limited perception, respectively.}

\Cref{tab:soa} provides a qualitative overview of the main approaches toward monitoring, contrasting robots in terms of traversability and resilience.
More specifically, traversability refers to the ability of the robot to traverse rough and complex environments, while resilience refers to the \rev{capacity to} resist and overcome unexpected events (falls, impacts, etc.).
Legged robots stand out in their potential to handle harsh, uneven landscapes such as scree habitats, where traditional modes of transport face significant obstacles.

\rev{It is important to highlight that we propose quadrupedal robots as an additional tool for habitat monitoring, rather than a full replacement for traditional methods or other effective robotic solutions such as drone-based remote monitoring.
The scientific community has already recognized the importance of supplementing existing methods, rather than replacing them, with robotic solutions~\cite{pringle2025opportunities}.
Quadrupedal robots are particularly well-suited for in-situ monitoring, where they can safely operate in close proximity to the monitored habitat, collect high-resolution data, and adapt to the same guidelines and protocols used by human botanists.
This makes them a valuable complement to existing monitoring practices, enhancing the capabilities of human botanists without fully replacing them.
}

\newcommand{\rating}[4]{%
    \begin{tikzpicture}[baseline=-0.6ex]
        \foreach \i in {1,...,4} {
            \ifnum\i>#1
                \filldraw[draw=black,fill=white] (0.25*\i-0.25,0) circle (0.8mm);
            \else
                \filldraw[draw=black,fill=black] (0.25*\i-0.25,0) circle (0.8mm);
            \fi
        }
    \end{tikzpicture}%
}

\begin{table}[htb]
    \renewcommand{\arraystretch}{1.3}
    \caption{Qualitative overview of robotic monitoring platforms.}\label{tab:soa}
    \centering
	\resizebox{.5\textwidth}{!}{%
    \begin{tabular}{lcccc}
        \toprule
        \textbf{Type} & \textbf{Traverasibility} & \textbf{Autonomy} & \textbf{Resilience} & \textbf{Others} \\
        \midrule
        Drones & \rating{4}{0}{0}{0} & \rating{1}{0}{0}{0} & \rating{1}{0}{0}{0} & Weather-dependant \\
        Wheeled & \rating{1}{0}{0}{0} & \rating{3}{0}{0}{0} & \rating{1}{0}{0}{0} & None \\
        SlothBot & \rating{2}{0}{0}{0} & \rating{4}{0}{0}{0} & \rating{1}{0}{0}{0} & Needs infrastructure \\
        \textbf{Quadrupeds} & \rating{3}{0}{0}{0} & \rating{2}{0}{0}{0} & \rating{3}{0}{0}{0} & None \\
        \bottomrule
    \end{tabular}
	}
\end{table}
\section{Scree Monitoring Approach}\label{sec:scree_mon}

\begin{figure}[tb]
    \centering
    \subfloat[\label{fig:plot_1_delimited}]
        {\includegraphics[width=.48\textwidth]{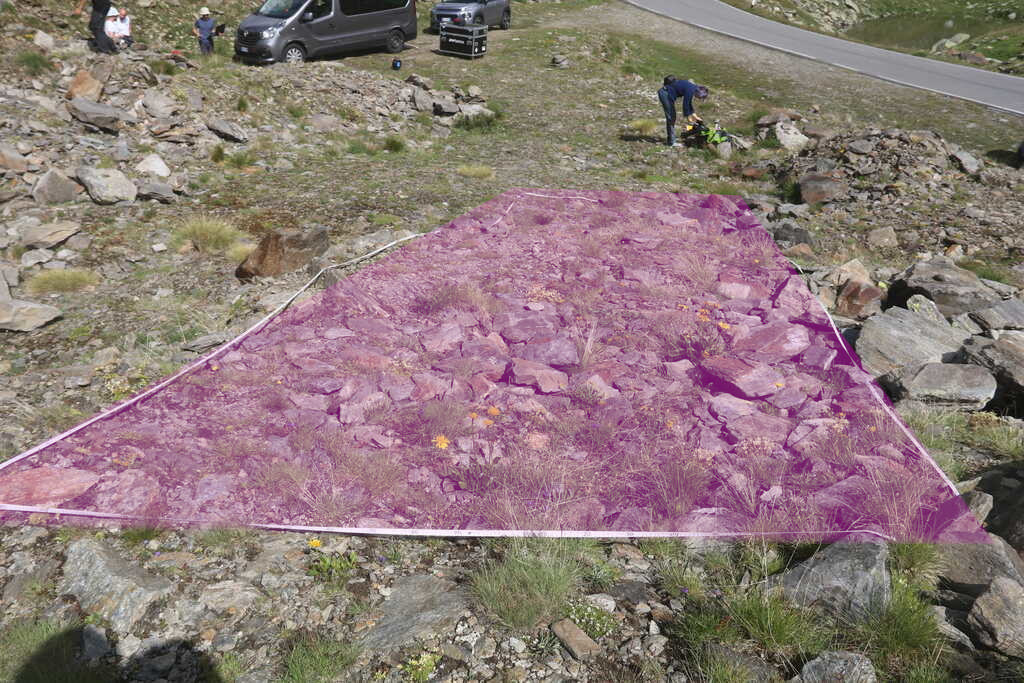}}\\
    \subfloat[\label{fig:plot_2_delimited}]{
        \includegraphics[width=.48\textwidth]{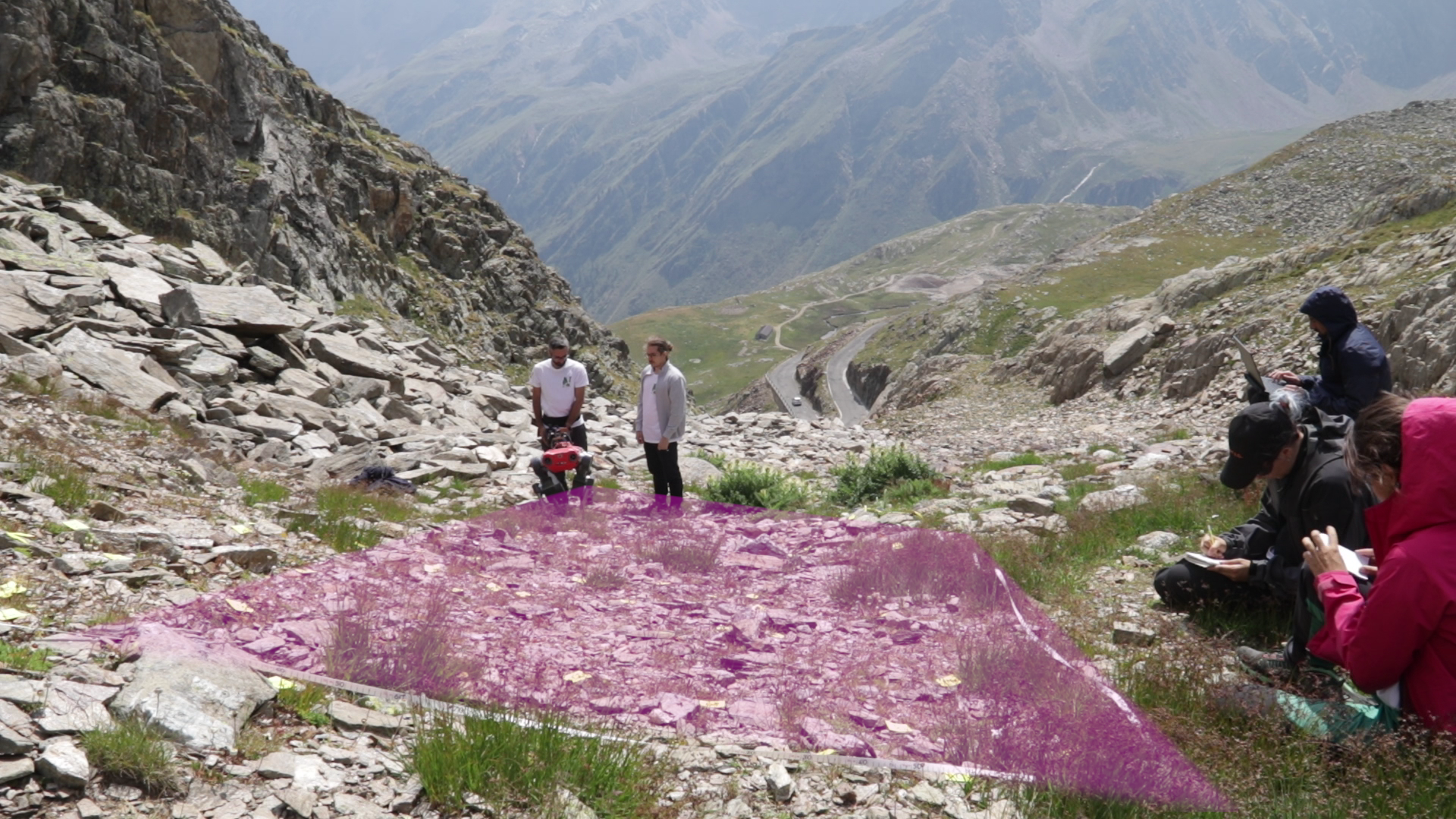}}
    \caption{Two scree plots manually delimited using a white tape. The monitoring area is highlighted in pink.
    (a) Plot NI 7. (b) Plot NI 2.}
    \label{fig:plot_delimited}
\end{figure}

\begin{figure*}[tb]
    \centering
    \resizebox{\textwidth}{!}{%
        \subfloat[\label{fig:stelvio_map}]
            {\includegraphics[height=6cm]{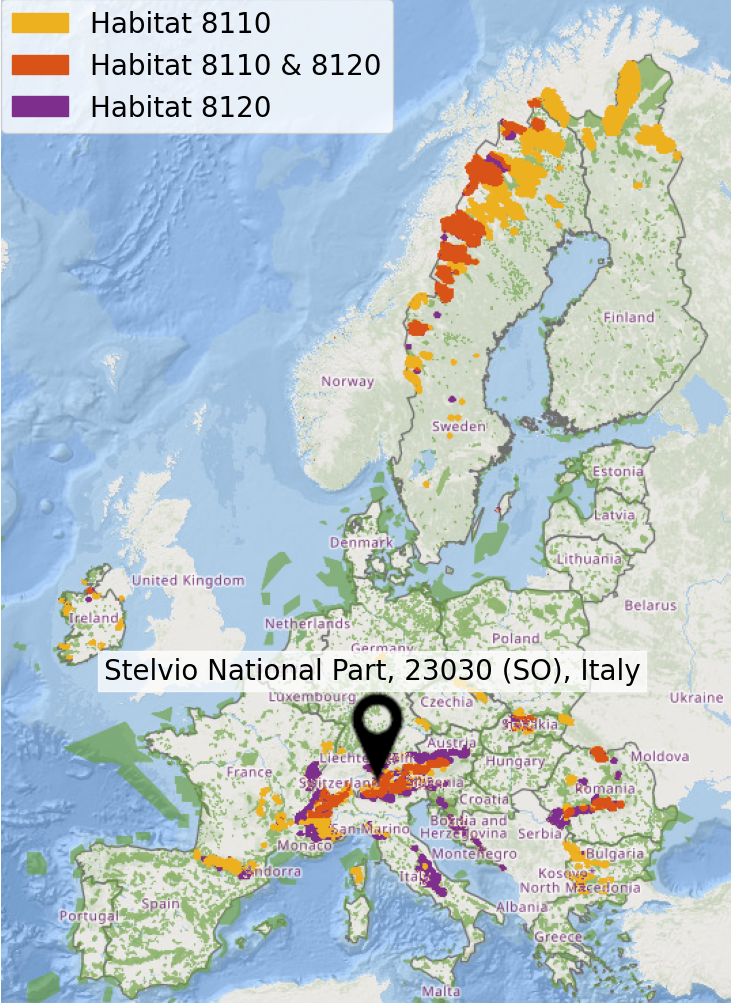}} \,
        \subfloat[\label{fig:mission_path_1}]
            {\includegraphics[height=6cm]{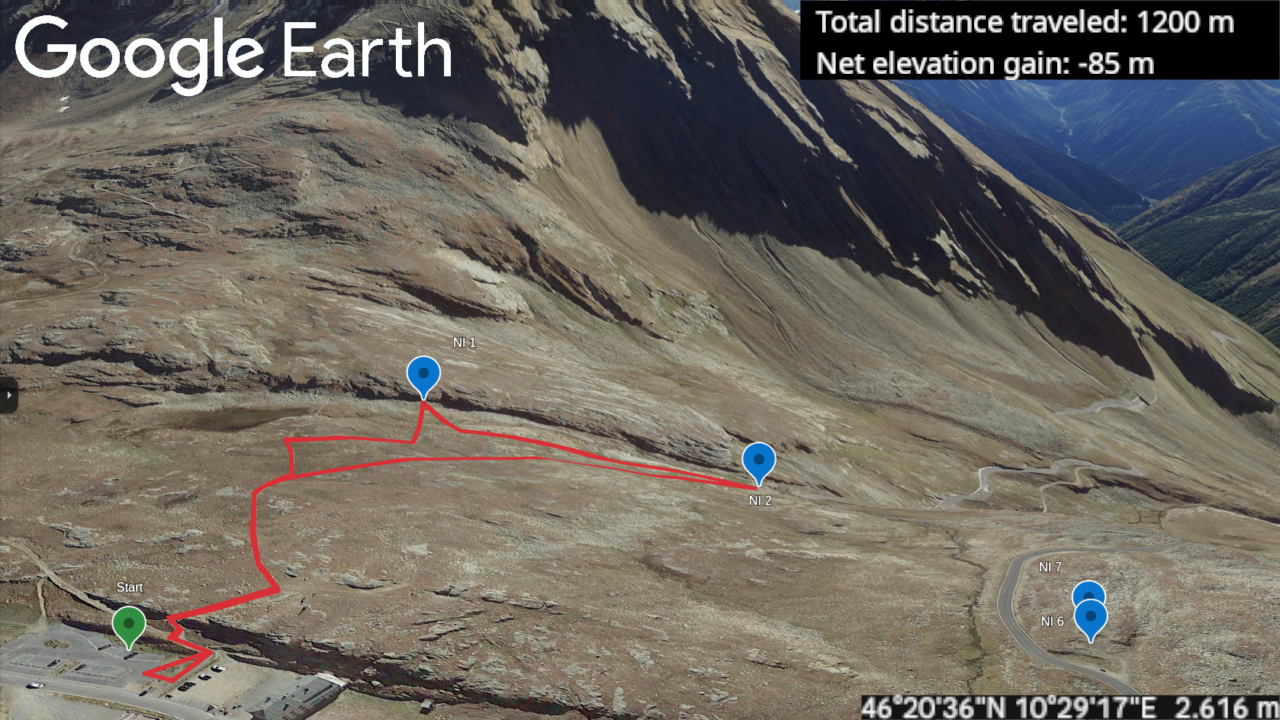}}
    }
    \resizebox{\textwidth}{!}{
        \subfloat[\label{fig:mission_path_2}]
            {\includegraphics[height=6cm]{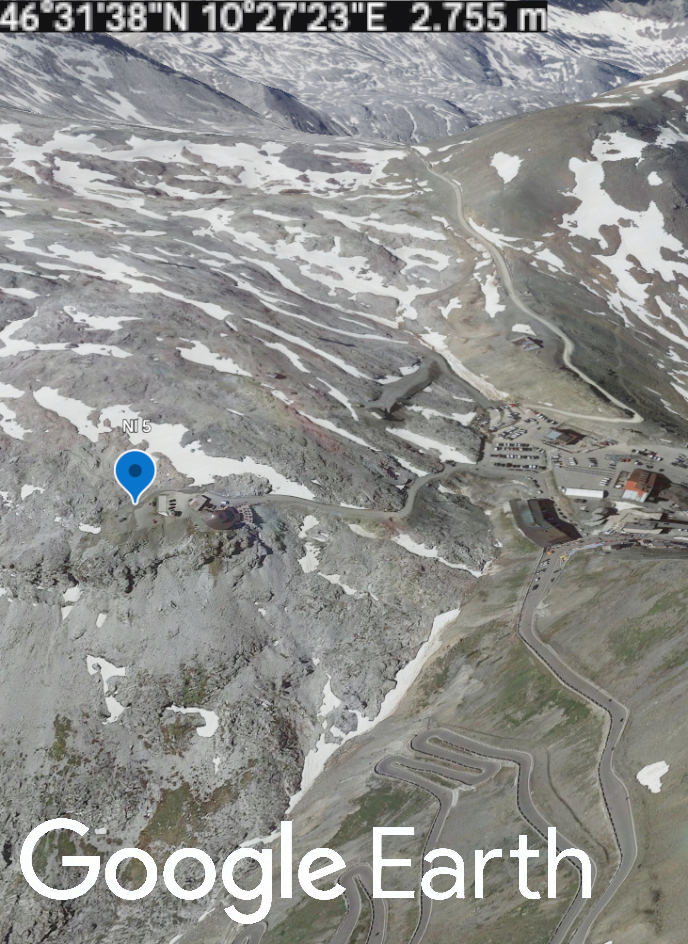}} \,
        \subfloat[\label{fig:mission_path_3}]
            {\includegraphics[height=6cm]{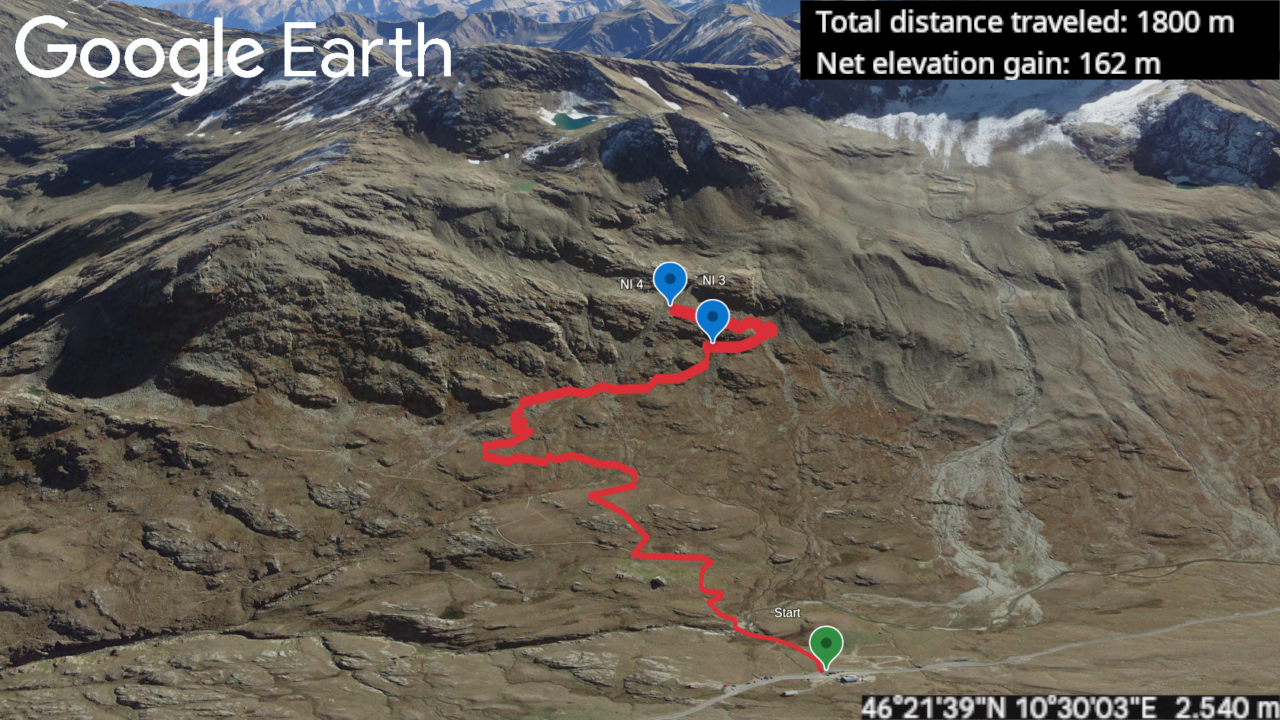}}
    }
    \caption{
        (a) Distributions of the habitats 8110 and 8120 in Europe, and approximate position of the field missions.
        (b) Path followed by ANYmal on the 19-th of July 2023. The starting point is represented with the green position mark, while the plot locations are represented with blue ones. On the right, the positions of the plots NI 6 and NI 7, explored on the 13-th July 2023, are shown.
        (c) Position of the NI 5 plot, explored on the 11-th July 2023.
        (d) Path followed by ANYmal on the 20-th of July 2023, during which the plots NI 3 and NI 4 were explored.
    }
    \label{fig:map_1}
\end{figure*}

\begin{table*}[tb]
    \centering
    \caption{Monitored plots of the two field campaigns with their relevant characteristics.\label{tab:monitored_plots}}
    \begin{tabular}[b]{lccccccc}
        \toprule
        \textbf{Plot name} & \textbf{Date} & \textbf{Time} & \textbf{Weather} & \textbf{Coordinates (N, E)} & \textbf{Habitat ID} & \textbf{Plot size} & \textbf{Max terrain slope} \\
        \midrule
        NI 1 & 19/07/22 & 13:00 & Partly cloudy & 46.3418444, 10.4909639 & 8110 & $4 \times \SI{4}{m}$ & $\SI{20}{\degree}$ \\
        NI 2 & 19/07/22 & 15:08 & Partly cloudy & 46.3409444, 10.4901170 & 8110 & $7 \times \SI{2}{m}$ & $\SI{30}{\degree}$ \\
        NI 3 & 20/07/22 & 14:04 & Partly cloudy & 46.3562722, 10.5086167 & 8110 & $4 \times \SI{3}{m}$ & $\SI{20}{\degree}$ \\
        NI 4 & 20/07/22 & 16:12 & Partly cloudy & 46.3561056, 10.5098333 & 8110 & $4 \times \SI{4}{m}$ & $\SI{25}{\degree}$ \\
        \midrule
        NI 5 & 11/07/23 & 10:52 & Partly cloudy & 46.5272980, 10.4565870 & 8120 & $3 \times \SI{9}{m}$ & $\SI{20}{\degree}$ \\
        NI 2 & 12/07/23 & 15:45 & Cloudy        & 46.3409444, 10.4901170 & 8110 & $7 \times \SI{2}{m}$ & $\SI{30}{\degree}$ \\
        NI 6 & 13/07/23 & 10:21 & Partly cloudy & 46.3413130, 10.4850260 & 8110 & $9 \times \SI{2}{m}$ & $\SI{20}{\degree}$ \\
        NI 7 & 13/07/23 & 11:26 & Cloudy        & 46.3411440, 10.4851220 & 8110 & $5 \times \SI{2}{m}$ & $\SI{20}{\degree}$ \\
        \bottomrule
    \end{tabular}
\end{table*}

This section presents the proposed robotic monitoring approach for alpine scree, along with a description of the robotic platform used.
First, \cref{sec:trad_scree_mon}, outlines the traditional monitoring methods typically employed by botanists for scree monitoring.
Then, \cref{sec:situ_desc}, introduces the specific Natura 2000 habitats addressed in this work.
Finally, \cref{sec:key_indicators} describes the key plant species used to evaluate scree conservation status.

\subsection{Traditional Scree Monitoring}\label{sec:trad_scree_mon}

In the EU, habitat monitoring is regulated according to the Habitats Directive (Art. 17 of the 92/43/EEC), which has become the pillar of biodiversity conservation.
Following the Habitats Directive, a habitat's conservation status is assessed by evaluating four parameters: (1) range, (2) area, (3) structure and functions, and (4) future prospects~\cite{ellwanger2018current}.
Among them, structure and functions is a key parameter.
However, monitoring is carried out heterogeneously in EU countries.
For instance, while the Habitats Directive recommends estimating the structure and functions parameter by analyzing the TS and EWS of a habitat, in Italy and other countries its evaluation is carried out by performing phytosociological relevées.
This type of relevée has the merit of providing more accurate and extensive information, at the cost of being more time-intensive and complex to perform.

Monitoring takes place in multiple spots, chosen following stratified sampling principles~\cite{elzinga1998measuring}, called plots.
The standardized scree plot has a rectangular shape and a minimum size of $\SI{16}{m^2}$.
Initially, the plot location (latitude, longitude, and altitude) is recorded, and its area is manually delimited.
Two instances of manually delimited plots are represented in~\cref{fig:plot_delimited}, where the plot regions are highlighted in pink.
Within each plot, botanists identify all plant species, estimate their ground cover, and note additional attributes such as blooming stages.  
Data is usually recorded on paper or via digital tools and subsequently used to assess the habitat's conservation status.

\subsection{Site Description}\label{sec:situ_desc}

Our field investigations focused on two Annex I habitats under the Habitats Directive:  
\begin{itemize}
    \item \textbf{Habitat 8110}: Siliceous scree of the montane to snow levels (Androsacetalia alpinae and Galeopsietalia ladani),
    \item \textbf{Habitat 8120}: Calcareous and calcshist scree of the montane to alpine levels (Thlaspietea rotundifolii).
\end{itemize}
Both habitats are classified as \emph{unfavourable-inadequate} (U1) in the fourth EU report~\cite{hab_dir_art_17}, signifying the need for more rigorous monitoring and conservation measures.
These high-altitude scree ecosystems occur in mountainous regions and are characterized by sparse, specialized plant communities that endure harsh climates and frequent substrate disturbances.
A map of the occurrence of these habitats in the Natura 2000 network is represented in~\cref{fig:stelvio_map}.

Two separate field campaigns were conducted in Valfurva (Sondrio, Italy) in July 2022 and July 2023, within the Stelvio National Park (SPA IT2040044) (see~\cref{fig:stelvio_map}).
More specifically, the first campaign took place between the \nth{18} and the \nth{22} of July 2022, while the second one between the \nth{10} and the \nth{14} of July 2023.
\rev{Botanists selected these dates to capture the full blooming of vascular plants of these habitats, which happens approximately during July and August, and to take advantage of good weather conditions and of the absence of snow \rev{\cite{valle2022biodiversity}}.}
\Cref{fig:mission_path_1,fig:mission_path_2,fig:mission_path_3} show the approximate paths followed by the robot during the two field campaigns and the positions of the monitored plots.
\Cref{tab:monitored_plots} details the monitored plots, including date and time, weather conditions, geographic coordinates, habitat identifier, plot size, and approximate maximum terrain slope.

\subsection{A New Protocol for Robotic Monitoring of Scree Habitat}\label{sec:key_indicators}

According to the Habitats Directive, Typical Species are essential for determining the conservation status of a habitat, yet the directive itself does not prescribe a standardized list nor even a rigorous definition of the term~\cite{bonari2021shedding}.
Instead, researchers have suggested some criteria for designating TS and EWS~\cite{gigante2016methodological}.
In Italy, many botanists rely on the Italian Interpretation Manual of the Habitats Directive~\cite{biondi2010manuale} to select pertinent TS and EWS for their studies.
Therefore, we followed the same criteria to select the TS and EWS for our robotic monitoring framework.
Given the challenges of identifying certain species exclusively from images, we grouped morphologically similar taxa, such as \emph{Cerastium uniflorum} Clairv. and \emph{Cerastium pedunculatum} Gaudin, under a single genus label \emph{Cerastium}.

For this research, the TS identified are:
\begin{itemize}
    \item \emph{Cerastium} spp.
    \item \emph{Geum reptans} L.
    \item \emph{Papaver alpinum} L.
    \item \emph{Ranunculus glacialis} L.
    \item \emph{Saxifraga bryoides} L. T.
\end{itemize}
The sole EWS included is \emph{Luzula alpinopilosa} Chaix Breistr.

Datasets including unlabeled and labeled images of these TS and EWS are available in~\cite{angelini2023alpine} and \cite{dilorenzo2025robotic}, respectively.

\section{Robotic Monitoring}\label{sec:rob_mon}

\begin{table}[htb]
	\caption{ANYmal C datasheet.}\label{tab:datasheet}
	\centering
	\begin{tabular}{ll}
		\toprule
        \textbf{Data} & \textbf{Value} \\
        \midrule
		Size Lying $\text{L} \times \text{W} \times \text{H}$ & $1054 \times 630 \times \SI{376}{mm}$ \\
        Size Standing, $\text{L} \times \text{W} \times \text{H}$ & $1054 \times 520 \times \SI{830}{mm}$ \\
        Weight & $\SI{50}{kg}$ \\
        Ingress Protection Rating & IP67 \\
        Operating Temperature Range & $0 - \SI{40}{\celsius}$ \\
        Maximum walking speed & $\SI{1}{m/s}$ \\
        Battery autonomy & $2 / \SI{3}{h}$ \\
		\bottomrule
	\end{tabular}
\end{table}

\begin{figure*}[tbh]
    \centering
    \includegraphics[width=.8\textwidth]{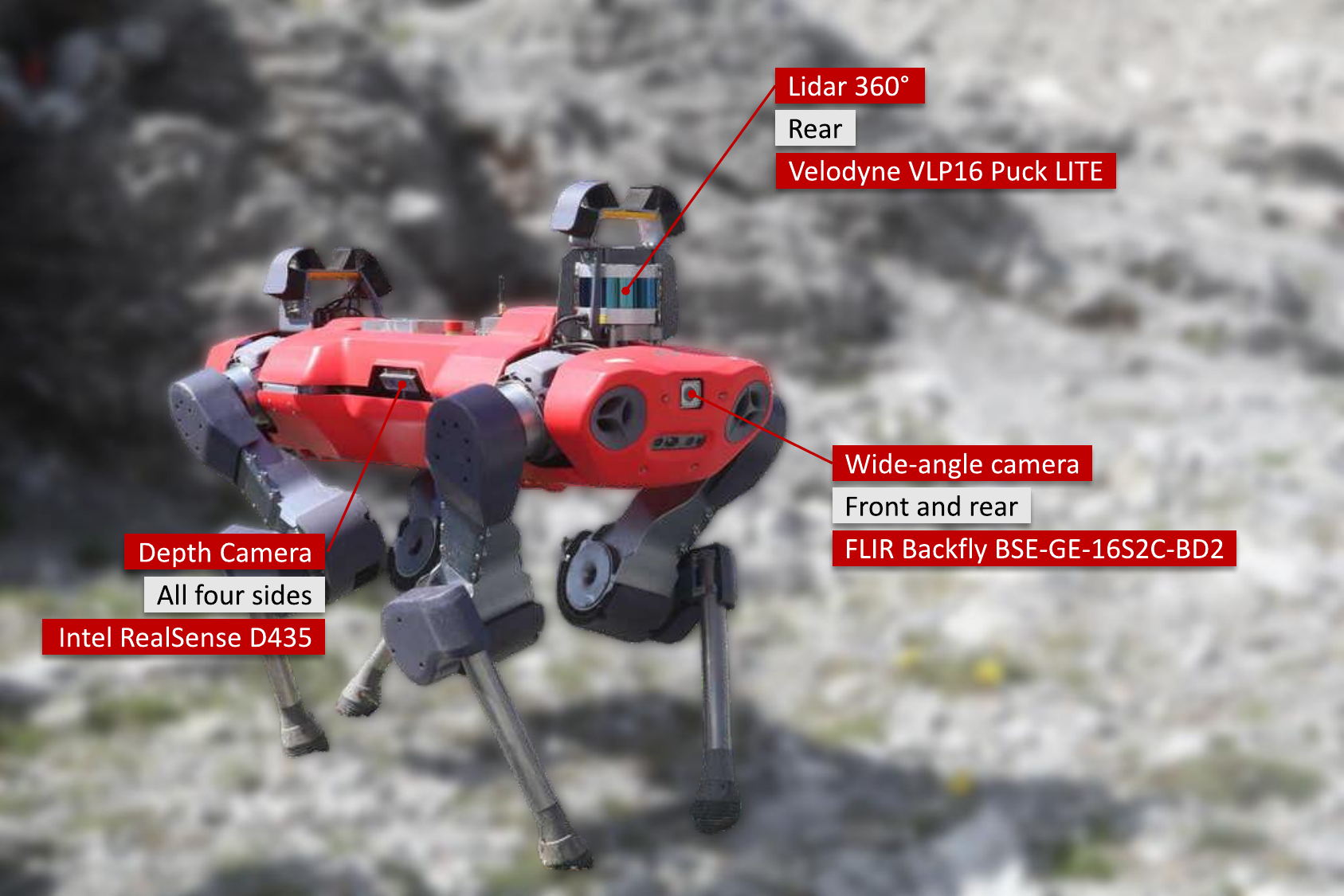}
    \caption{Overview of the exteroceptive sensors equipped on the ANYmal C robot.}
    \label{fig:anymal_sensors}
\end{figure*}

This section presents the robot deployed in the field campaigns and the framework established for robotic monitoring.
More specifically, \cref{sec:rob_equip} provides an overview of the robotic platform used in the field campaigns, highlighting its capabilities and sensing equipment.
\Cref{sec:rob_mon_mission} details the framework established for robotic monitoring.
Lastly, \cref{sec:slippage_metric,sec:veg_cov_est,sec:detection_nn} discuss the three primary components of the robotic monitoring process: the slippage metric, vegetation cover estimation, and detection neural network.

\subsection{Robotic Equipment}\label{sec:rob_equip}

Among commercial quadrupeds, ANYmal C stands out as a robust and agile platform, well-suited for challenging environments~\cite{angelini2023robotic}.
Additionally, a fully open research version is available for higher freedom of software customization.
\Cref{tab:datasheet} summarizes key specifications.

The actuation system of ANYmal C is based on Series Elastic Actuators (SEAs).
Its legs each have three joints powered by SEAs, which confer the robot with structural softness.
This characteristic has proven to be fundamental for efficient and robust locomotion and is fundamental for locomoting in an extremely challenging environment, such as the scree habitat. 

ANYmal C is also equipped with a wide array of sensors for localization and monitoring purposes (\cref{fig:anymal_sensors}).
These include proprioceptive (IMU and joint encoders) and exteroceptive sensors, such as a lidar for SLAM-based localization, a GPS unit for outdoor applications, and multiple RGB-depth and wide-angle cameras.  
These cameras are central to habitat monitoring, enabling the collection of high-resolution visual data for plant identification and coverage estimation. 

The robot can either be teleoperated or used autonomously.
When teleoperated, a human operator equipped with a remote controller moves the \rev{robot through horizontal linear velocity and yaw rate commands}.
Conversely, during autonomous missions, the robot fuses lidar-based mapping and proprioceptive localization to localize itself and follow predefined waypoint paths.

\rev{Additionally, the robot has a removable battery, which allows for quick battery swaps in the field, ensuring minimal downtime during missions.
Each battery provides approximately two to three hours of autonomy and has a charging time from empty to full charge (with $\SI{8}{A}$) of three hours.
It has dimensions of $466 \times 136 \times \SI{78.1}{mm}$ and weighs $\SI{5.55}{kg}$.}

\begin{figure*}[htb]
    \centering
    \resizebox{\textwidth}{!}{%
        \subfloat[\label{fig:nn_spatial}]
            {\includegraphics[height=6cm]{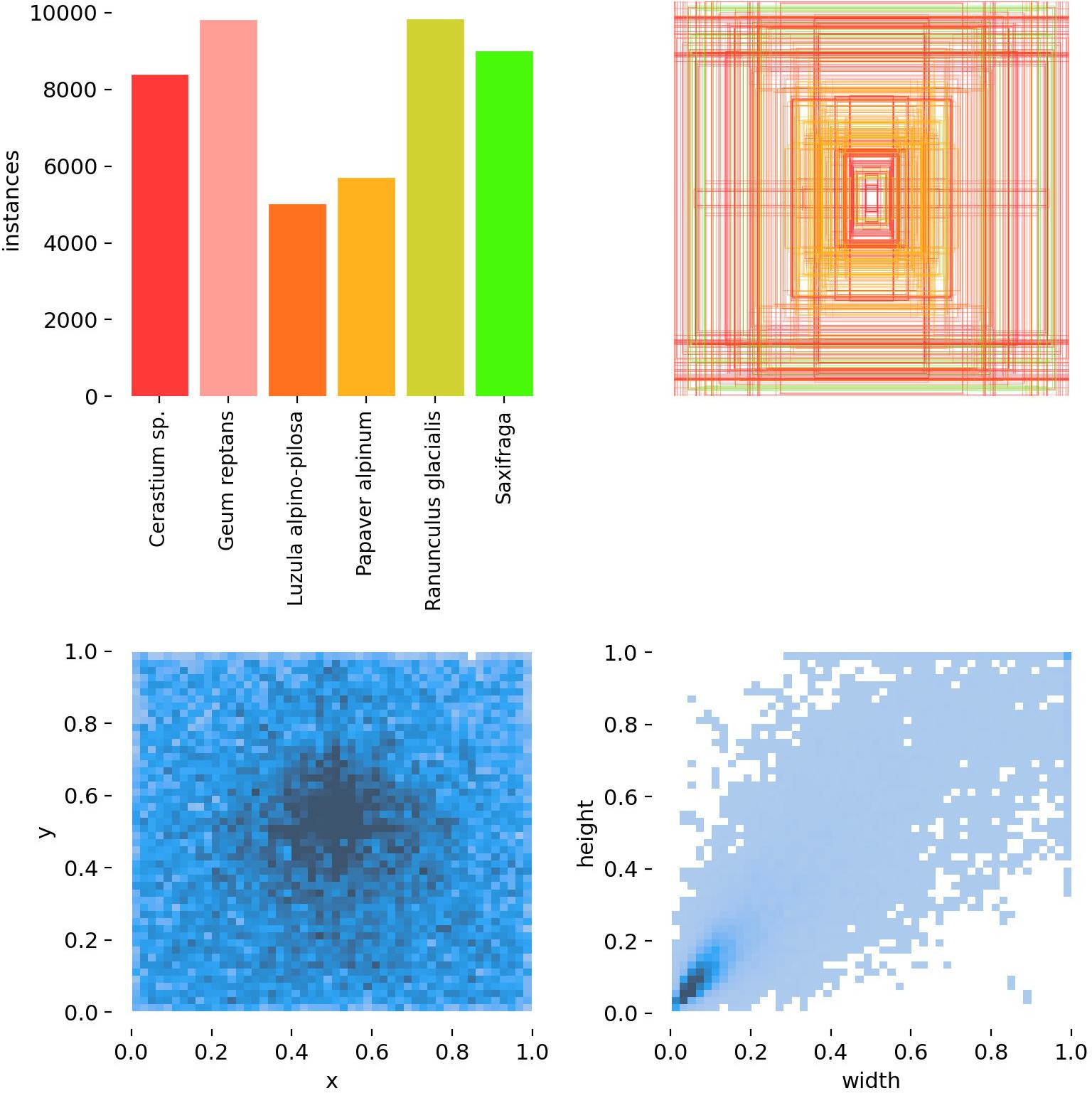}} \quad
        \subfloat[\label{fig:nn_correlogram}]
            {\includegraphics[height=6cm]{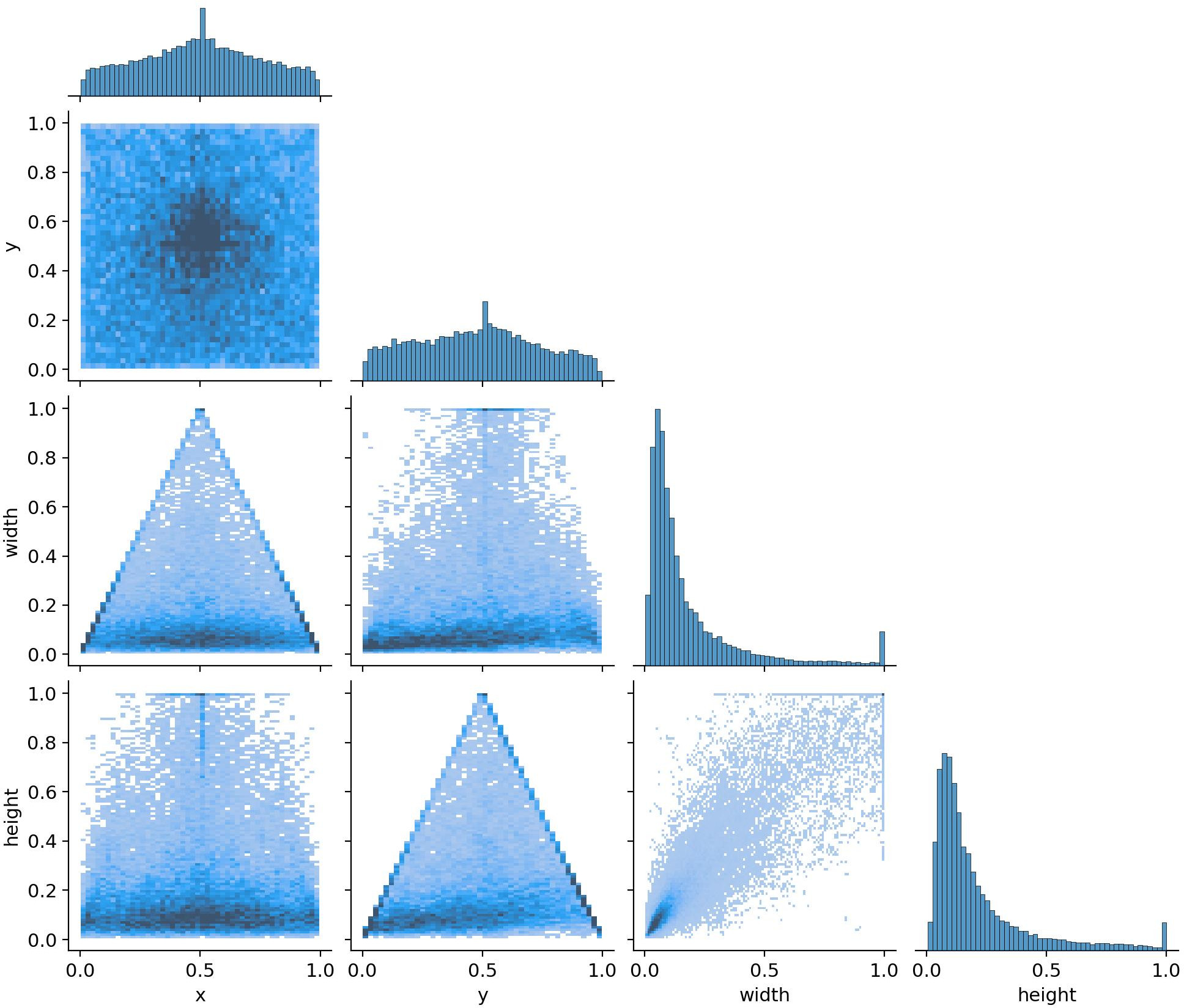}}
    }
    \caption{\rev{(a) Spatial and class distribution of the plant labels. From left to right, top to bottom: the class frequencies histogram, the intersection heatmap of annotated bounding boxes (highlighting areas of spatial label density and overlap), 2D density plots of centroid coordinates, and the bounding-box shapes. (b) Correlogram of the normalized YOLO labels: $x$ and $y$ are the coordinates of the bounding-box center (relative to image width and height, respectively); width and height represent the bounding-box dimensions, also normalized by image size.}}
    \label{fig:nn_dataset}
\end{figure*}


\subsection{Robotic Monitoring Description}\label{sec:rob_mon_mission}

Our proposed robotic protocol parallels traditional scree monitoring workflows to ensure data comparability.
It consists of two main stages: \emph{mapping} and \emph{autonomous survey}.

\textbf{Mapping:} This phase is required by the robot to achieve accurate localization during the monitoring mission.
An operator familiar with the robot\rev{'}s software conducts a brief mapping run to generate a 3D point cloud of the area.  
During this phase, it can be advantageous to move the quadruped base to acquire new data points from a different perspective and increase the map quality.
The localization accuracy depends on the point density and the features provided by the environment.
For scree sites, a mapping duration of about a minute or less, and little to no repositioning of the robot proved to be sufficient to build a reliable map.

\textbf{Autonomous Survey:}
After mapping, a grid-based mission plan is defined, covering the rectangular region of interest.
Waypoints located at the grid centers guide the robot\rev{'}s navigation, ensuring comprehensive coverage of the plot from multiple viewpoints.  
The robot begins at a designated start point outside the grid, traverses each waypoint (pausing briefly for data capture), and returns to the start point.  
During the mission, the onboard cameras continuously record video, and at each waypoint, synchronized still images are captured.  
These high-resolution images form the basis for subsequent plant identification and habitat analysis, \rev{which is performed offline, after the mission, as online processing would require additional computational resources with no added benefit.}

\rev{
The waypoint trajectory is designed to ensure a good trade-off between data quality (e.g., sufficient overlap between images) and autonomous mission duration.
Waypoints spaced at $\SI{1.0}{m}$ intervals combine adequate data coverage with a reasonable mission duration.
The problem of optimally traversing a grid of waypoints is referred to as the traveling salesman problem (TSP), and is generally NP-hard.
However, with a square grid and under the reasonable assumption that all the points in the grid are reachable and that no obstacles are present, an optimal hamiltonian path (i.e., a path that visits each point exactly once) is trivially obtained by visiting the waypoints in a serpentine manner (see~\cref{fig:mission_grid}).
}

\rev{
\textbf{Rationale for Autonomy:}
We opt for a fully autonomous survey because it (1) shortens mission time by removing operator reaction delays, (2) guarantees repeatability of the monitoring process, and (3) improves time efficiency, since the botanist can focus on other activities.
These gains in efficiency, repeatability, and labor usage are essential for scaling scree monitoring.
}
\subsection{Slippage Metric}\label{sec:slippage_metric}

The slippage is an important factor in evaluating the locomotion capabilities of legged robots.
Therefore, estimating it has been investigated even in the absence of direct contact sensors, as is the case with the ANYmal robot.
Following~\cite{camurri2020pronto}, the probability of a foot being in (stable) contact is modeled through a sigmoid function that depends on the estimated contact force.
\begin{equation}
    P(c_\mathrm{i} = 1 | \vec{f}_i) = \frac{1}{1 + \exp(- \beta_1 f_{z,i} - \beta_0)}, \qquad i = 1, \dots, 4
\end{equation}
where $c_\mathrm{i}$ is the contact state of the $i$th foot (1 if in contact, 0 otherwise), $\vec{f}_i$ is the $i$th estimated contact force, $f_{z,i}$ is the vertical component of $\vec{f}_i$, and $\beta_0$ and $\beta_1$ are the sigmoid parameters learned using a logistic classifier~\cite{camurri2020pronto}.

The estimated contact forces are obtained from the inverse dynamics of the robot as follows
\begin{equation}
    \vec{f} = - \left( \vec{J}(\vec{q})^T \right)^{\dagger} \left( \vec{\tau} - \vec{h} - \vec{f}_\text{s}^T \begin{bmatrix} \dot{\vec{v}} \\ \dot{\vec{\omega}} \end{bmatrix}  \right)
\end{equation}
where $\vec{J}(\vec{q})$ is the contacts Jacobian matrix, $\vec{q}$ the generalized coordinates vector, $\vec{\tau}$ the joint torques, $\vec{h}$ the Coriolis and centrifugal forces, $\vec{f}_\text{s}$ the spatial forces at the floating base, $\dot{\vec{v}}$ the linear acceleration of the base, $\dot{\vec{\omega}}$ the angular acceleration of the base, and $\square^{\dagger}$ the Moore-Penrose pseudoinverse.

Given the robot's feet contact state ($c_\mathrm{i}$), it is possible to compute the slippage metric as in~\cite{catalano2021adaptive} with
\begin{equation}
    s = \frac{\sum_{i = 1}^4 \int_{t_0}^{t_\mathrm{f}} c_\mathrm{i}(t) v_\mathrm{i}(t) dt}{\int_{t_0}^{t_\mathrm{f}} v_\mathrm{b}(t) dt}
\end{equation}
where $v_\mathrm{i}$ is the velocity of the $i$th foot, $v_\mathrm{b}$ the velocity of the base, $t$ is the time, and $t_0$ and $t_\mathrm{f}$ the initial and final times.

\subsection{Vegetation Cover Estimation}\label{sec:veg_cov_est}

\begin{figure*}[htb]
    \centering
    \subfloat[\label{fig:hab_small_stones}]
        {\includegraphics[width=.3\textwidth]{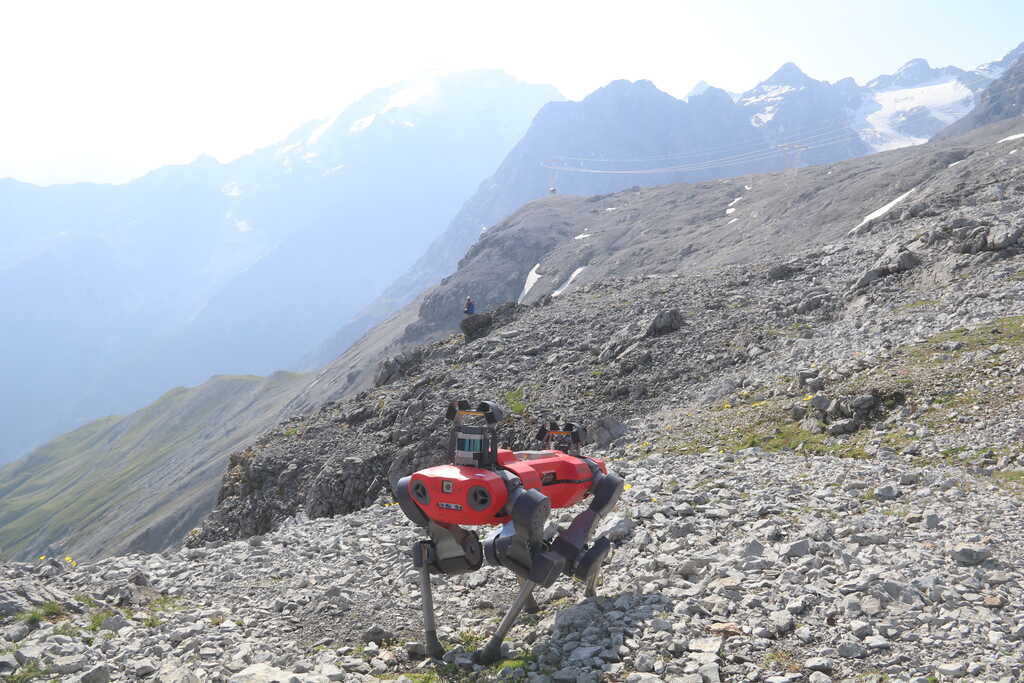}} \,
    \subfloat[\label{fig:hab_unstable_boulders}]
        {\includegraphics[width=.3\textwidth]{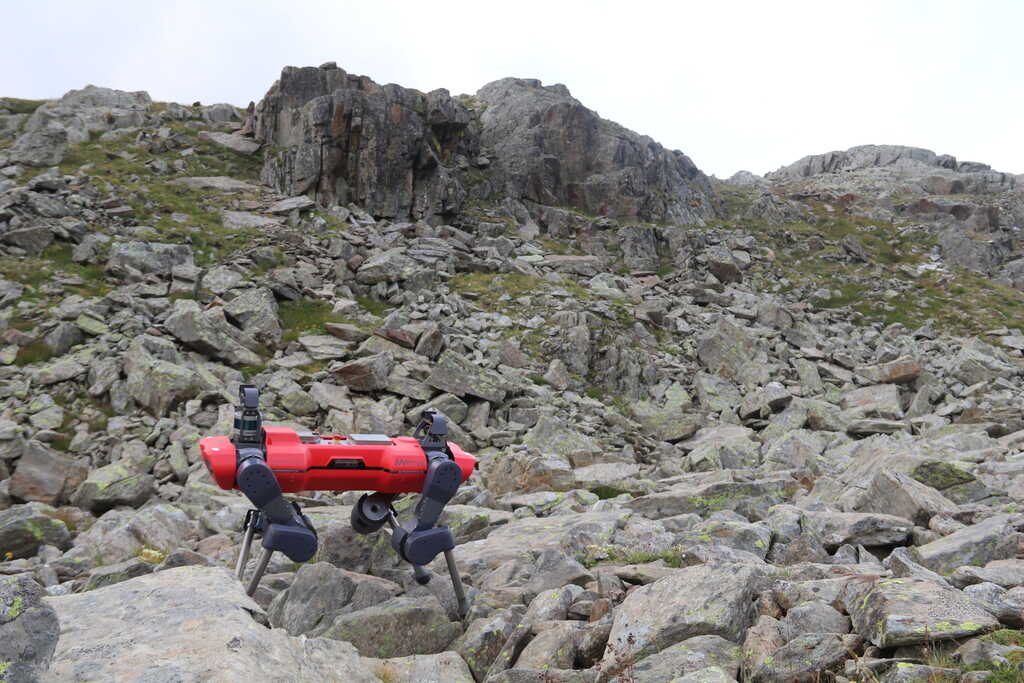}} \,
    \subfloat[\label{fig:hab_obstacles}]
        {\includegraphics[width=.3\textwidth]{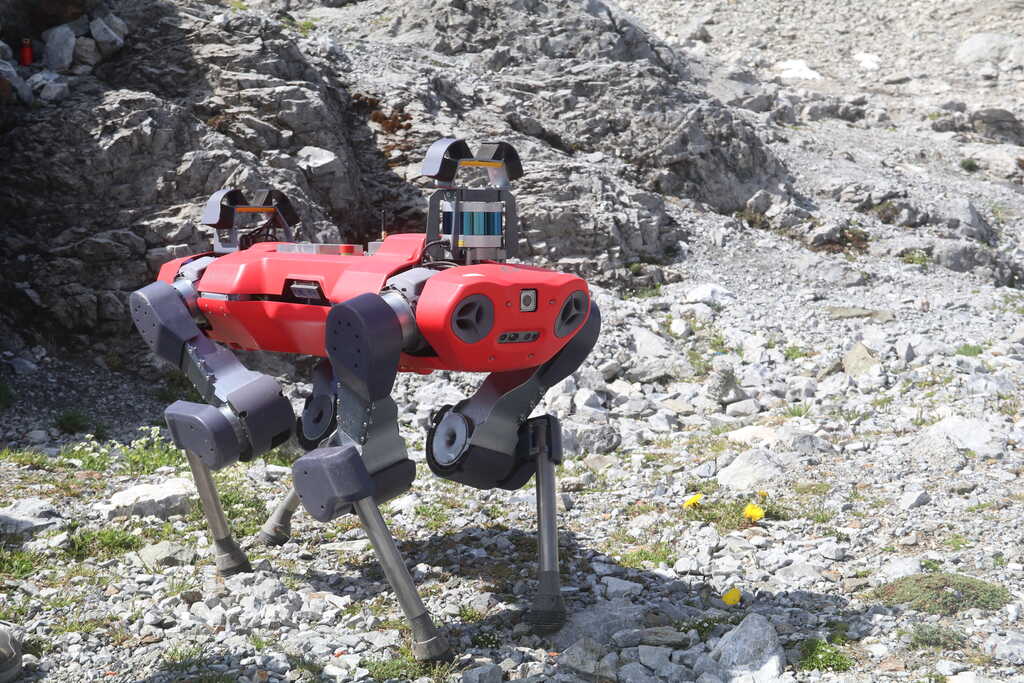}} \vspace{-10pt} \\
    \subfloat[\label{fig:hab_small_stones_2}]
        {\includegraphics[width=.3\textwidth]{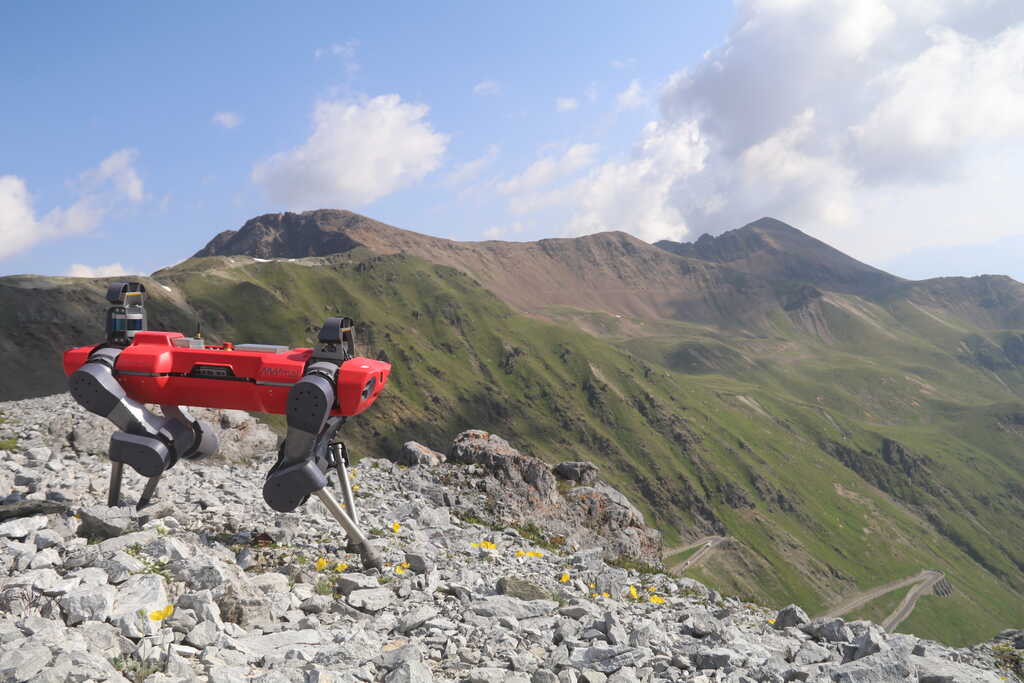}} \,
    \subfloat[\label{fig:hab_steep_inclination}]
        {\includegraphics[width=.3\textwidth]{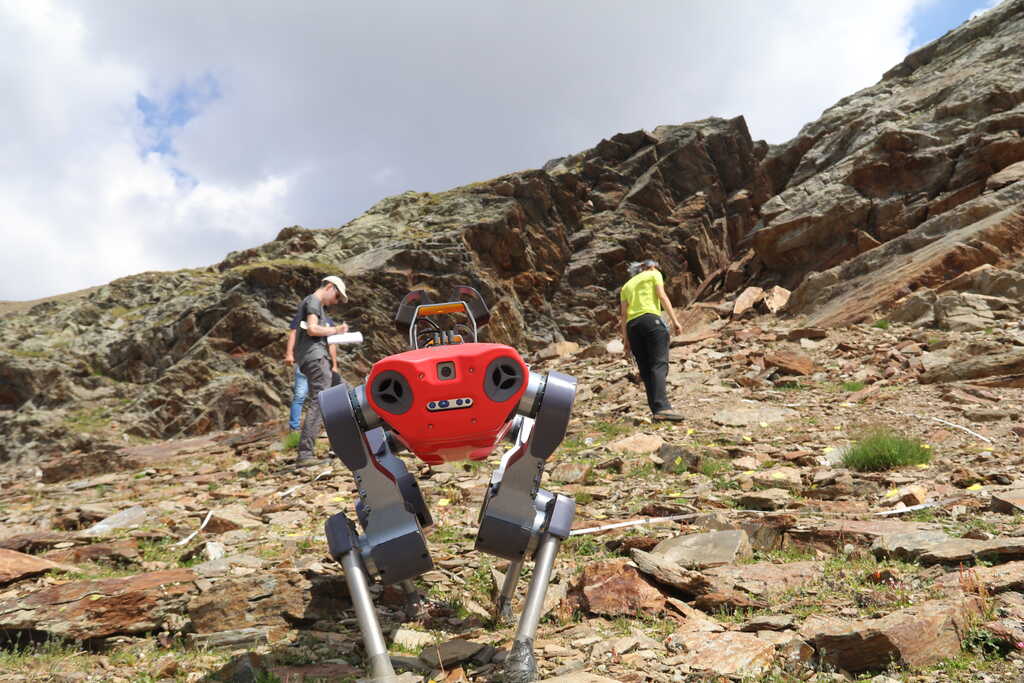}} \,
    \subfloat[\label{fig:hab_harsh_weather}]
        {\includegraphics[width=.3\textwidth]{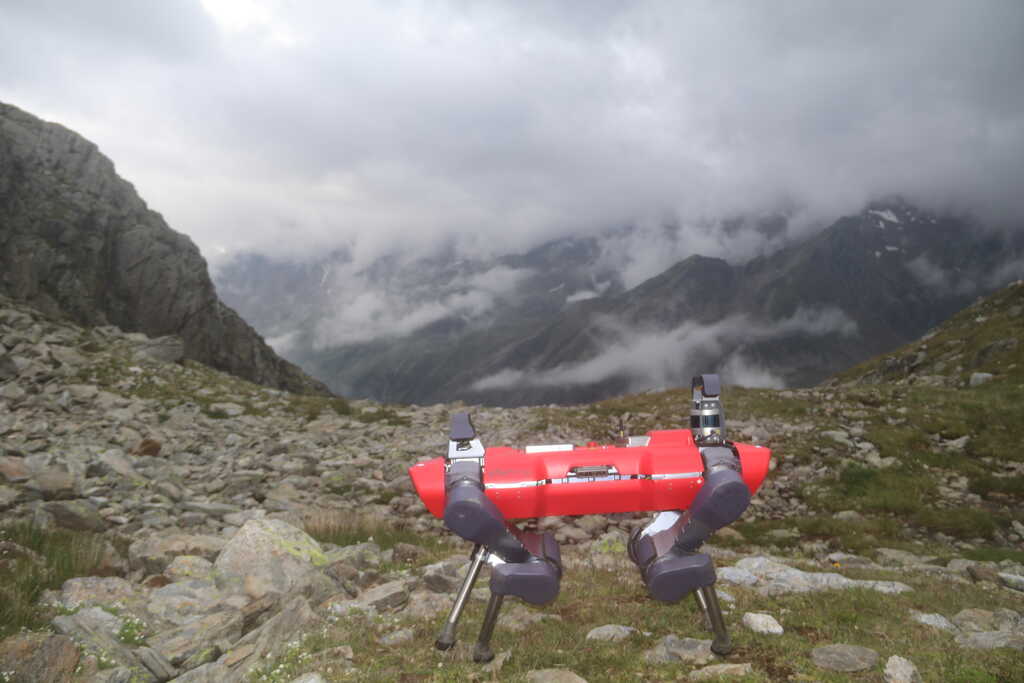}} \vspace{-10pt} \\
    \subfloat[\label{fig:hab_grass}]
        {\includegraphics[width=.3\textwidth]{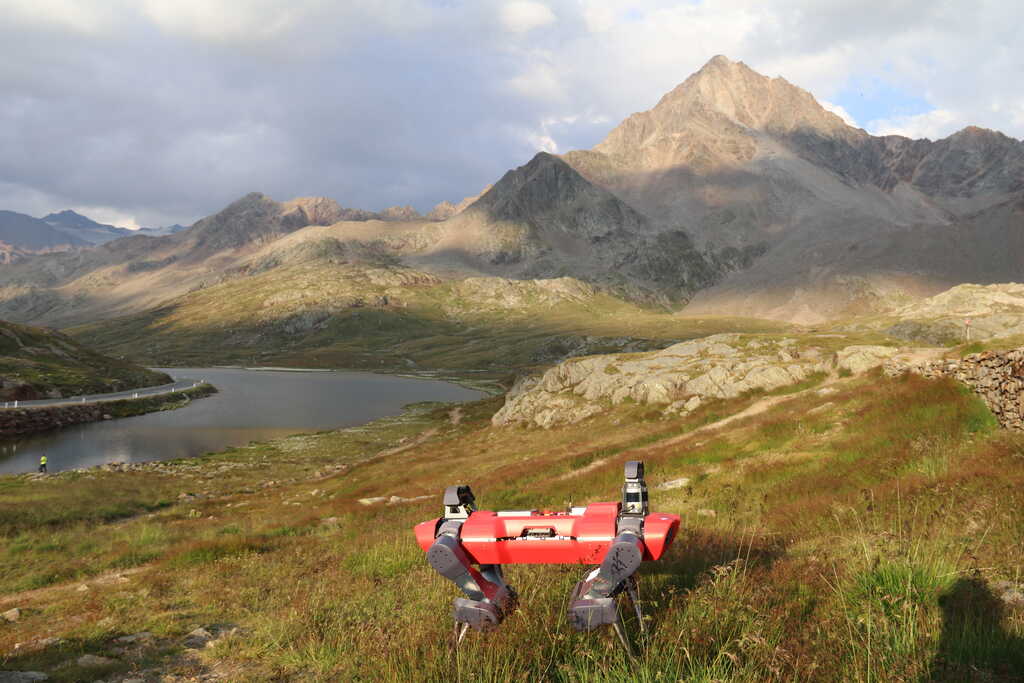}} \,
    \subfloat[\label{fig:hab_grass_2}]
        {\includegraphics[width=.3\textwidth]{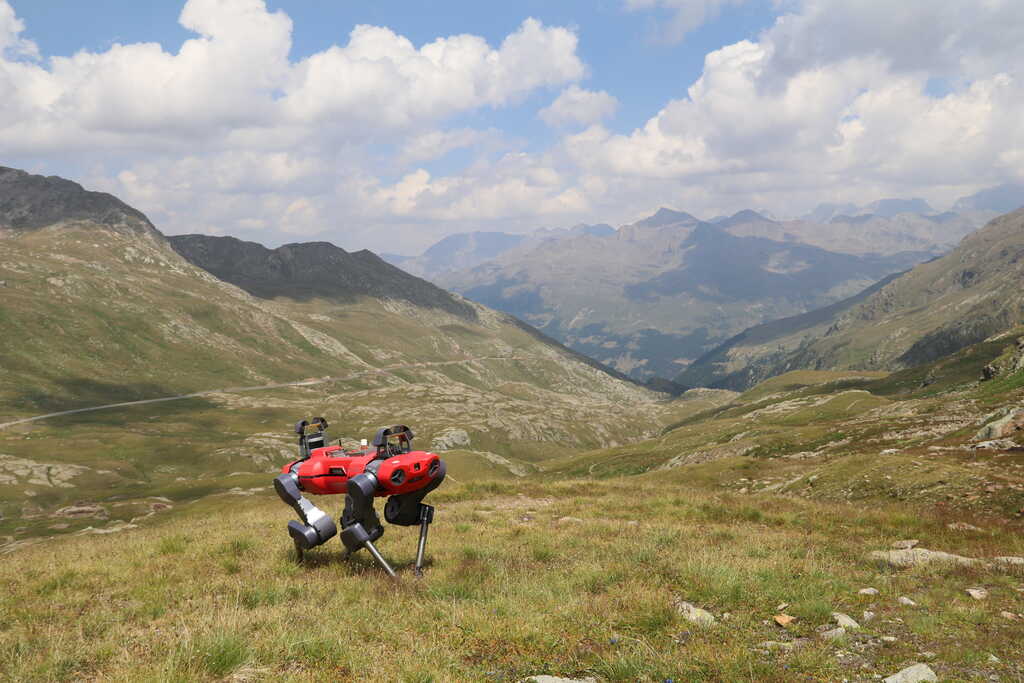}} \,
    \subfloat[\label{fig:hab_slippery_spot}]
        {\includegraphics[width=.3\textwidth]{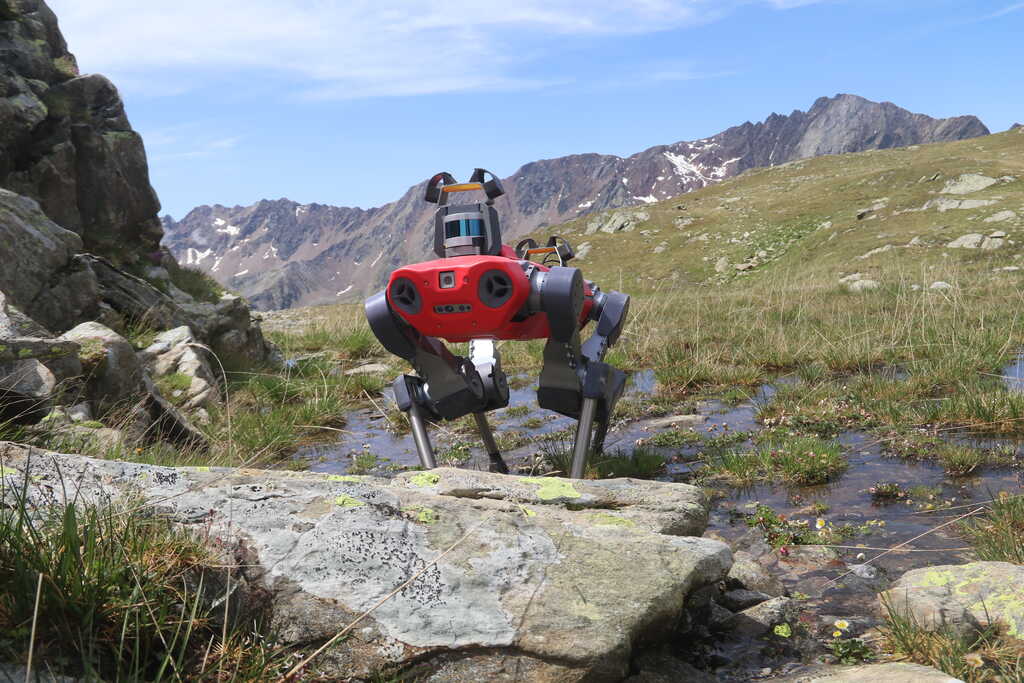}}
    \caption{
        Different terrains and weather conditions typical of the scree habitat and its surroundings.
        (a) Small stones. (b) Unstable boulders. (c) Obstacles. (d) Small stones. (e) Steep inclination. (f) Harsh weather conditions. (g), (h) Grass. (i) Slippery spot.
    }
    \label{fig:scree_terrains}
\end{figure*}

\begin{figure*}[htb]
    \centering
    \setlength\tabcolsep{2pt}%
    \begin{minipage}[c]{\textwidth}
        \centering
        \subfloat[\label{fig:mission_grid}]
            {\includegraphics[width=.5\textwidth]{img/mission_snapshots_1/1_grid}}
        \subfloat[\label{fig:mission_map}]
            {\includegraphics[width=.5\textwidth]{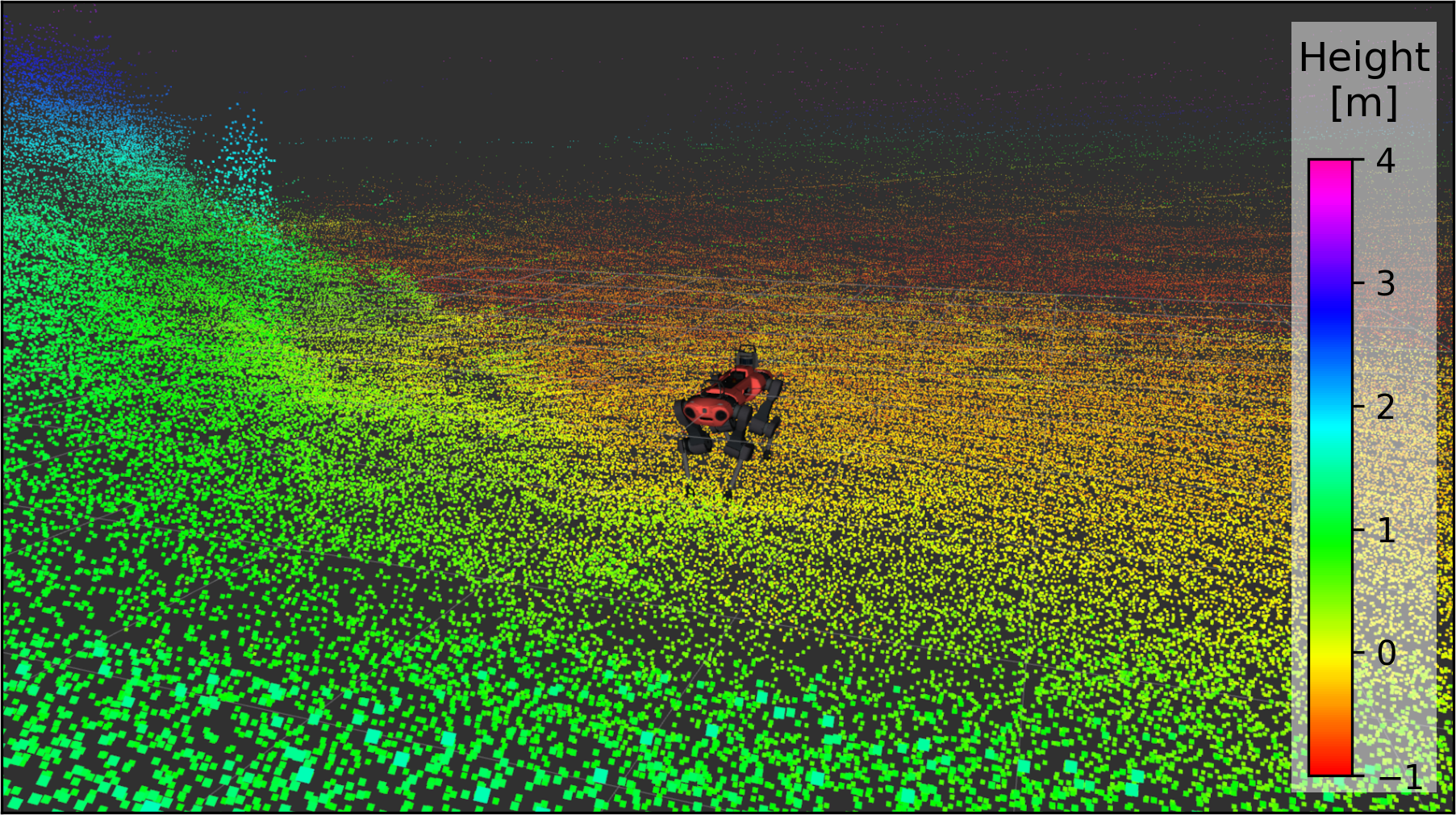}}
    \end{minipage} \\
    \begin{minipage}[c]{\textwidth}
        \centering
        \subfloat[\label{fig:mission_snapshots}]{%
                \includegraphics[width=\linewidth]{img/mission_snapshots_1/yolo_anymal}
        }
    \end{minipage}
    \caption{
        (a) NI 1 plot with the square grid (in orange) and the waypoint trajectory (in blue) superimposed on the image.
        (b) \rev{Point cloud of the surroundings obtained with the lidar during the mapping phase.} The color of the points indicates their height.
        (c) Overlapped snapshots and actual trajectory (in yellow) of ANYmal performing a monitoring mission. The corresponding video is available in the supplementary material.
    }
    \label{fig:mission}
\end{figure*}

\begin{figure*}[htb]
    \centering
    \subfloat[\label{fig:depth_front_0}]{\includegraphics*[width=0.235\textwidth]{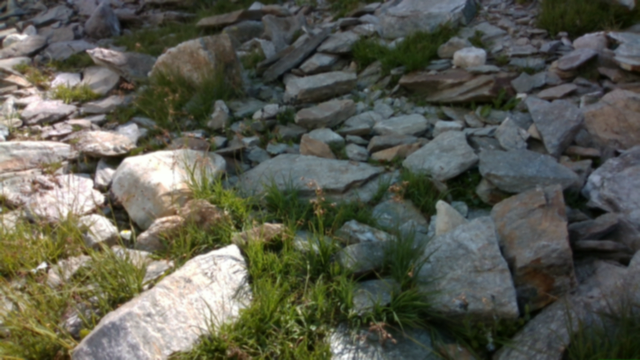}}\,
    \subfloat[\label{fig:depth_front_1}]{\includegraphics*[width=0.235\textwidth]{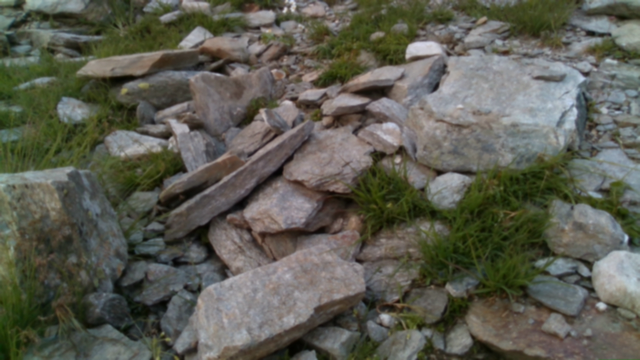}}\,
    \subfloat[\label{fig:depth_left_0}]{\includegraphics*[width=0.235\textwidth]{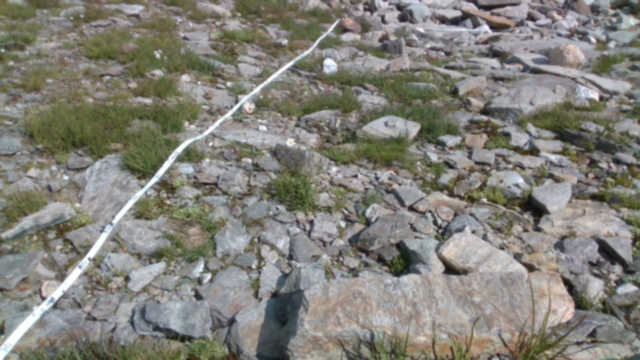}}\,
    \subfloat[\label{fig:depth_left_1}]{\includegraphics*[width=0.235\textwidth]{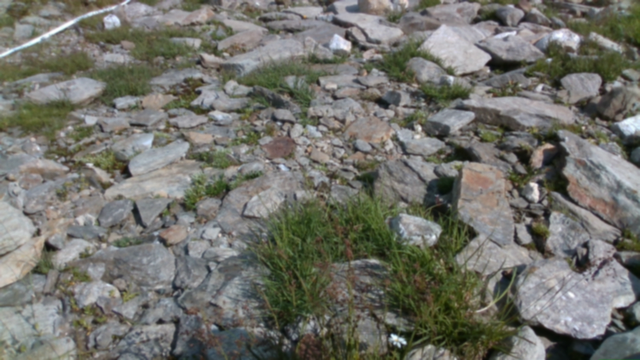}} \vspace{-10pt} \\
    \subfloat[\label{fig:depth_rear_0}]{\includegraphics*[width=0.235\textwidth]{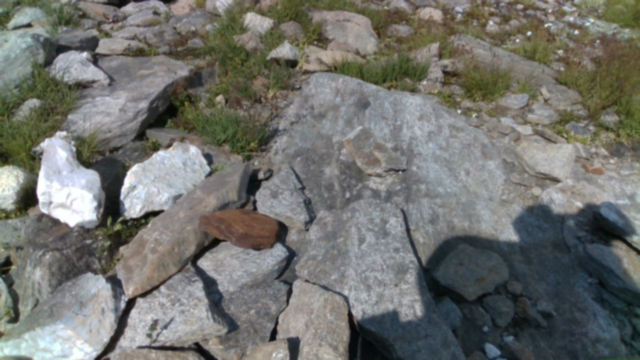}}\,
    \subfloat[\label{fig:depth_rear_1}]{\includegraphics*[width=0.235\textwidth]{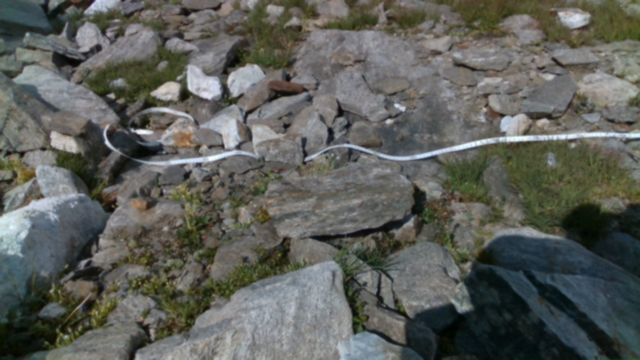}}\,
    \subfloat[\label{fig:depth_right_0}]{\includegraphics*[width=0.235\textwidth]{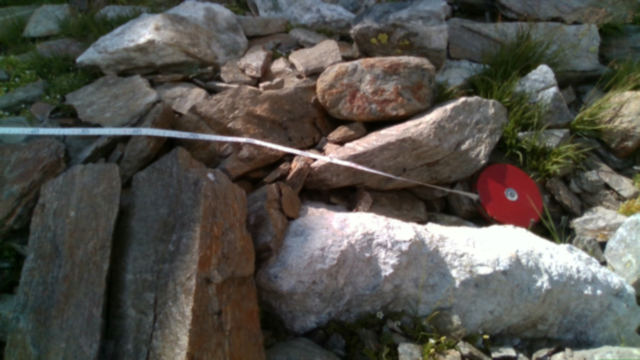}}\,
    \subfloat[\label{fig:depth_right_1}]{\includegraphics*[width=0.235\textwidth]{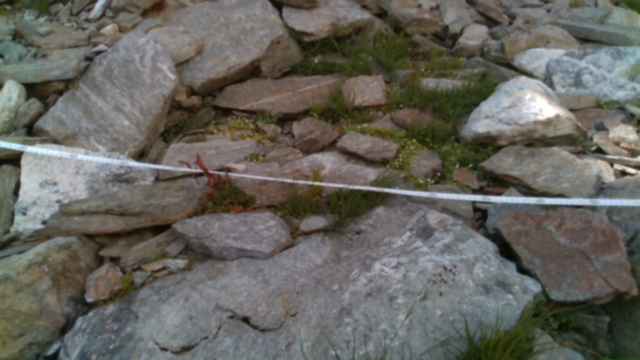}}
    \caption{\rev{Images} acquired by the \rev{four depth cameras} during the mission. \rev{The point of view is different from that of a standing botanist, being close to the ground while retaining a wide field of view.}}\label{fig:depth_images}
\end{figure*}

Several methods exist for estimating from images the vegetation cover percentage in a given area.
In this work, we employed the Excess Green Index (ExGI)~\cite{larrinaga2019greenness} for the vegetation cover estimation.
This index works by analyzing the red (R), green (G), and blue (B) channels of the image to distinguish vegetation from soil.
The ExGI is computed as follows
\begin{equation}
    \text{ExGI} = (2 \text{G}) - (\text{R} + \text{B}). \label{eq:exgi}
\end{equation}
When the index is greater than a certain threshold $t$, the corresponding pixel is considered as part of the vegetation.
Afterward, the vegetation cover percentage is obtained by computing the ratio of vegetation (according to the ExGI) in the whole image.

This index has been chosen because of its ease of use and because it has been proven that it outperforms other RGB-based indexes~\cite{larrinaga2019greenness}.

\subsection{Detection Neural Network}\label{sec:detection_nn}

The proposed robot-aided monitoring allows for the easy gathering of a lot of vegetation data.
More specifically, during the monitoring mission, the ANYmal robot acquires videos and photos of its surroundings from its set of cameras.
The data quality and quantity, coupled with the favorable point of view of the robot for observing the scree vegetation, make this data suitable for automating plant detection.

Considering the requirements and the objectives of the scree habitat monitoring, a neural network for object detection represents the optimal trade-off between capabilities and complexity~\cite{fernandez2023exploring}.
The developed network is based on YOLOv9~\cite{yolo}, a state-of-the-art model for classification, detection, segmentation, and more.
\rev{This choice is motivated by the fact that YOLOv9 is both an efficient and lightweight model and has demonstrated good performance even in challenging scenarios and with partially occluded objects.
Additionally, the potential real-time inference opens up the possibility of using the results and estimated accuracy of the detection to adapt the robot's path and improve the monitoring mission in future works.}

Due to the absence of datasets of the scree plant species in the existing literature, it was built during the two monitoring campaigns.
It consists of photos acquired by the robot cameras and of photos acquired by operators using a point of view similar to that of the robot.
Overall, it amounts to 2823 images.

A team of botanists drew a bounding box and labeled each plant of interest in the dataset.
The labeling was performed using the OpenLabelling tool, and the annotations were converted into the YOLOtxt format, suitable for training with YOLOv9.

\Cref{fig:nn_dataset} gives an overview of the dataset employed, showing the class distribution and the labels' correlogram.
\rev{The dataset includes 6 classes of plants with very heterogeneous sizes and dispositions in the camera field of view (see \cref{fig:nn_spatial,fig:nn_correlogram}).
The dataset is balanced, with a comparable number of annotated instances for each class, ensuring that the network is not biased toward any particular species.}

Before the training, the images were resized to $640 \times \SI{640}{pixels}$.
The proposed results were obtained by training the gelan-c architecture in Google Colab for 300 epochs, using an NVIDIA A100-SXM4-40GB GPU, starting from the dataset built internally.
Being small in size, a split train/validation/test was carried out, which sees 2186, 546, and 91 images, respectively.

\section{Results}\label{sec:results}

\begin{figure*}[htb]
    \centering
    \resizebox{\textwidth}{!}{%
        \subfloat[\label{fig:mon_times}]
            {\includegraphics[height=6.0cm]{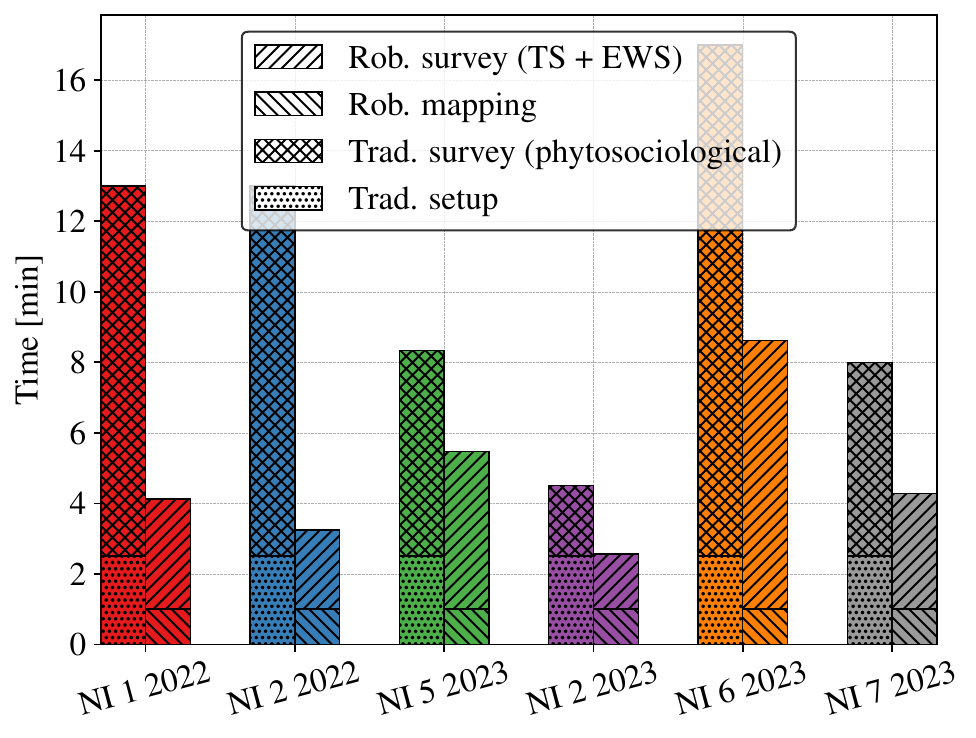}} \,
        \subfloat[\label{fig:mon_inclinations}]
            {\includegraphics[height=6.0cm]{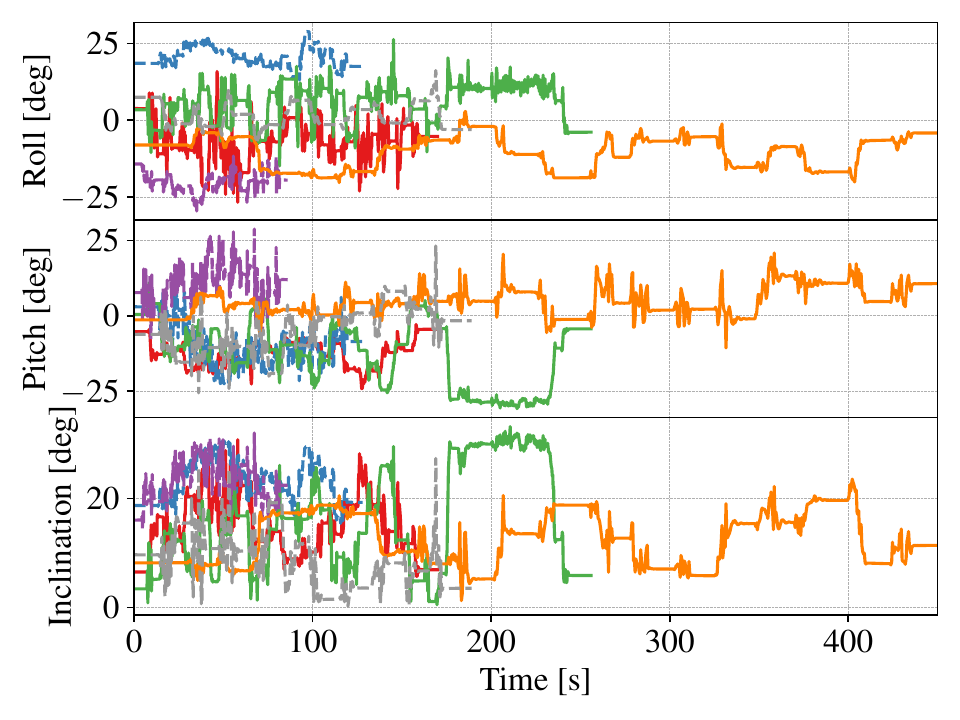}}
    }
    \resizebox{\textwidth}{!}{%
        \subfloat[\label{fig:mon_battery_usage}]
            {\includegraphics[height=4cm]{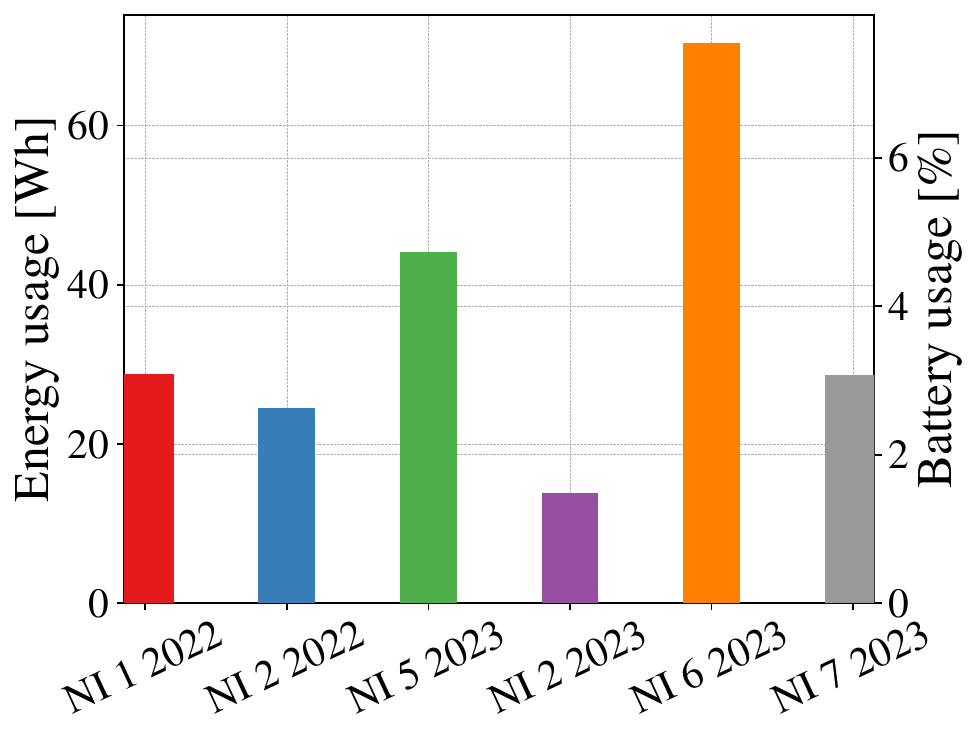}}\,
        \subfloat[\label{fig:mon_power_usage}]
            {\includegraphics[height=4cm]{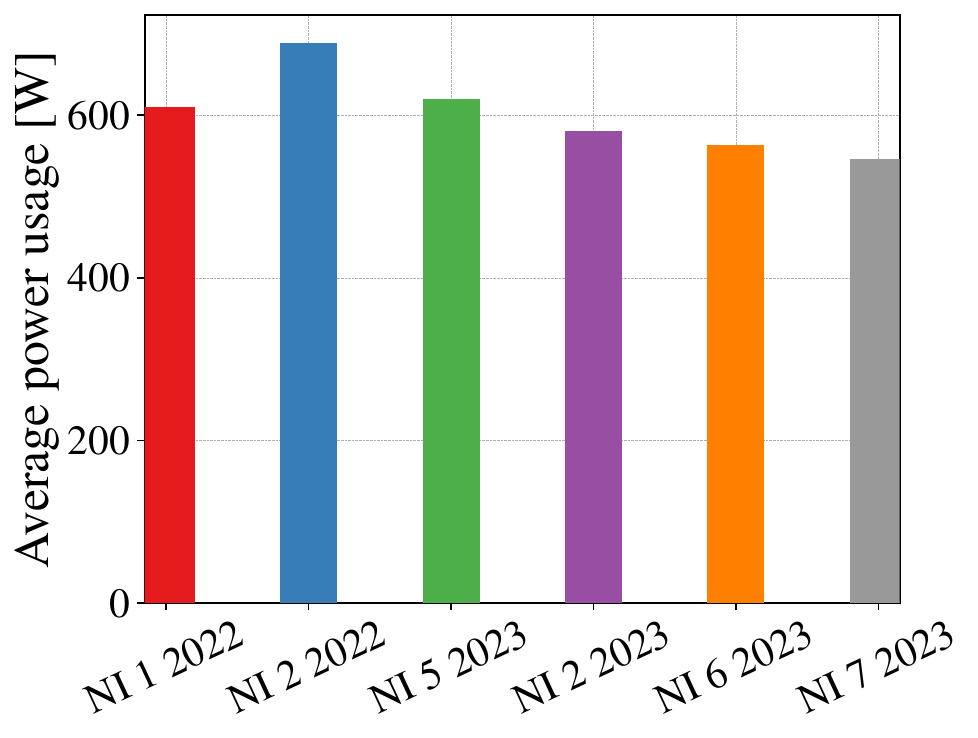}}\,
        \subfloat[\label{fig:mon_slippage}]
        {\includegraphics[height=4cm]{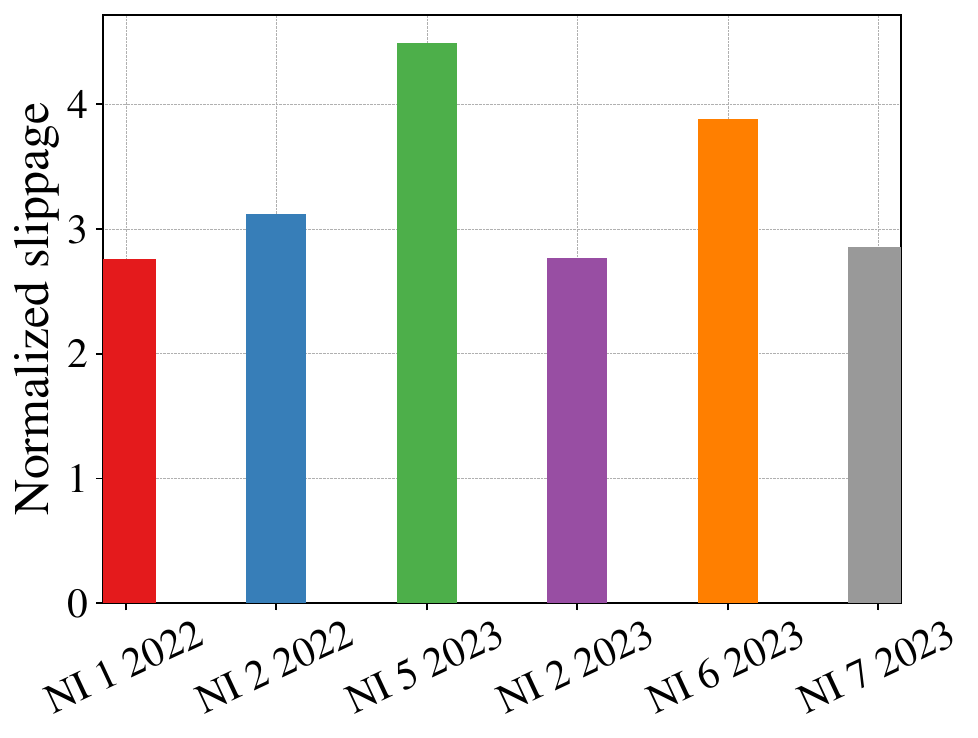}}
    }
    \caption{
        \rev{Robotic monitoring results. In all the plots, each line color represents a different monitoring mission.}
        (a) Monitoring times of the traditional (trad.) and robotic (rob.) monitoring. Note that the traditional monitoring consisted in the full phytosociological survey, while the robotic monitoring performed the detection of the TS and EWS and the vegetation cover estimation.
        (b) Roll and pitch inclinations and total inclination of the terrain during the monitoring missions.
        (c) Energy and battery usage during the monitoring missions.
        (d) Power usage during the monitoring missions.
        (e) Normalized slippages during the monitoring missions.
    }
    \label{fig:monitoring_results}
\end{figure*}

\begin{figure*}[htb]
    \centering
    \subfloat[]
        {\includegraphics[width=0.24\textwidth]{img/slip/1_diff}} \,
    \subfloat[]
        {\includegraphics[width=0.24\textwidth]{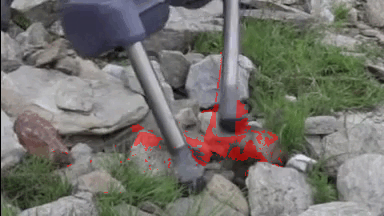}} \,
    \subfloat[]
        {\includegraphics[width=0.24\textwidth]{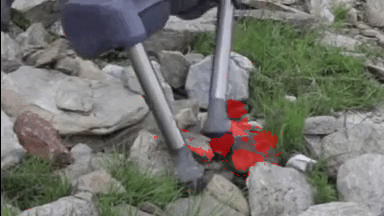}} \,
    \subfloat[]
        {\includegraphics[width=0.24\textwidth]{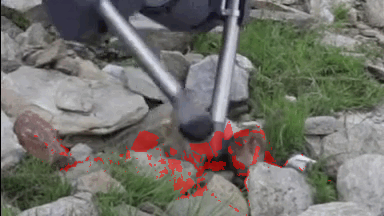}} \vspace{-10pt} \\
    \subfloat[]
        {\includegraphics[width=0.24\textwidth]{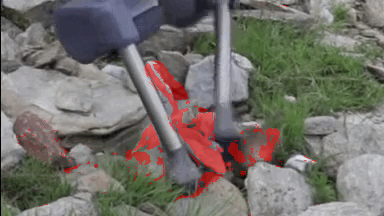}} \,
    \subfloat[]
        {\includegraphics[width=0.24\textwidth]{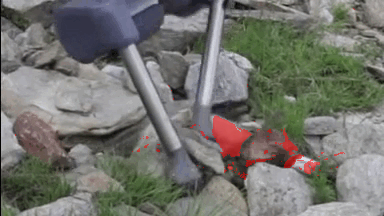}} \,
    \subfloat[]
        {\includegraphics[width=0.24\textwidth]{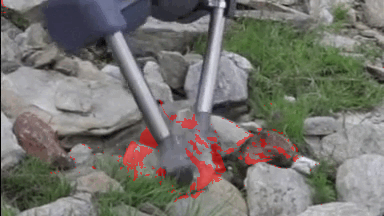}} \,
    \subfloat[]
        {\includegraphics[width=0.24\textwidth]{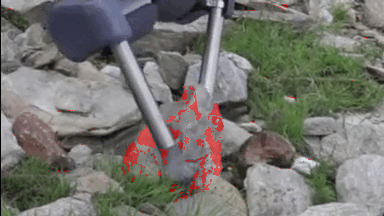}}
    \caption{Snapshots of the terrain moving under the robot's feet during the monitoring mission. \rev{The red overlay highlights the terrain movement between two consecutive frames caused by the robot's locomotion.}}
    \label{fig:slip_snap}
\end{figure*}

\subsection{Locomotion in the Scree Habitat}{\label{sec:loc_scree_hab}}

\rev{During the field campaigns, the required robotic equipment consisted of the ANYmal C robot, the remote controller, and one spare battery.
Additionally, a laptop was used to start the mapping and the autonomous monitoring missions.
All equipment was driven to the closest road-accessible point; from there, the operator piloted the robot to each survey site.
For maximum safety, the robot's speed was limited in software at $\SI{0.8}{m/s}$, approximately matching the botanists' walking pace and ensuring that ANYmal did not add appreciable delay to the missions.
As this was a pilot study, the equipment could be reduced by removing the laptop, and the robot hardware could be exploited to its full potential, allowing for considerably faster locomotion when the terrain conditions allow it.
}

One of the main challenges of monitoring a scree habitat stems from operating in such an unstructured and challenging environment, which is challenging even for humans.
A reliable locomotion ability is a fundamental requirement both for moving the robot to the desired monitoring location and for performing the actual monitoring.

\Cref{fig:scree_terrains} represents some terrain conditions in which the robot is required to operate during scree monitoring.
The main challenge is represented by the fact that scree is composed of rock fragments of different sizes and shapes (see~\cref{fig:hab_small_stones,fig:hab_unstable_boulders}), which can be unstable and move during locomotion.
Moreover, the gaps between the various rocks can lead to the robot's feet getting stuck or stumbling due to the change of the base of support (\cref{fig:hab_unstable_boulders}).
The habitat is also characterized by very steep inclines (greater than 40°) (\cref{fig:hab_steep_inclination}) and obstacles (\cref{fig:hab_obstacles}).
Additionally, scree surroundings can also be covered in grass or mud (\cref{fig:hab_grass,fig:hab_grass_2}), or slippery areas (\cref{fig:hab_slippery_spot}) may need to be traversed to reach the chosen monitoring location.
Furthermore, the habitat tends to be affected by harsh and unstable weather conditions such as fog, strong winds, snow, sleet, and rain (\cref{fig:hab_harsh_weather}).

Undeterred by these problems, the robot was able to \rev{reach} the designated monitoring spots efficiently and securely.
\rev{Two paths followed during the monitoring campaigns are represented in \cref{fig:mission_path_1,fig:mission_path_3}.
In these two days, the robot traveled approximately for $\SI{1.2}{km}$ and $\SI{1.8}{km}$, respectively, and had a net elevation change of $\SI{-85}{m}$ and $\SI{160}{m}$, respectively.}
It was decided against reaching the top of the Sforzellina's Glacier (at the \rev{very} top of \cref{fig:mission_path_3}), since the high slopes, the presence of narrow passages, and the slippery terrain due to snow and ice was very challenging and \rev{could} have posed \rev{a risk} to the robot's safety.
This level of performance and reliability was achieved using ANYmal's Trekker \rev{locomotion} controller, a blind controller based on reinforcement learning~\cite{lee2020learning}.

\subsection{The Efficiency of Robotic Monitoring}\label{sec:efficiency_rob_mon}

\definecolor{TP}{rgb}{0.275, 0.620, 0.208}  
\definecolor{TN}{rgb}{0.396, 0.263, 0.129}  
\definecolor{FP} {rgb}{1.000, 0.647, 0.000}  
\definecolor{FN} {rgb}{0.000, 1.000, 1.000}  

\begin{figure*}[htb]
    \centering
        \includegraphics[width=.24\textwidth]{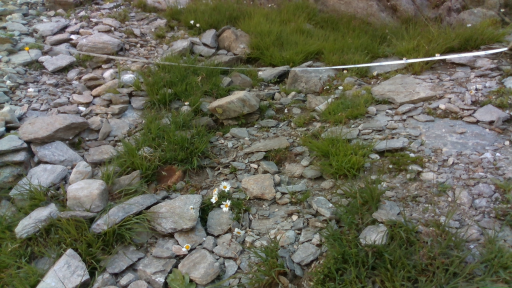}
        \includegraphics[width=.24\textwidth]{img/veg_cov/2.png} \,
        \includegraphics[width=.24\textwidth]{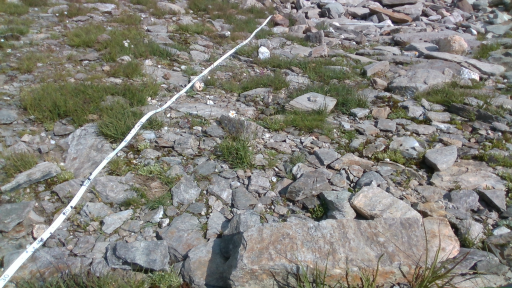}
        \includegraphics[width=.24\textwidth]{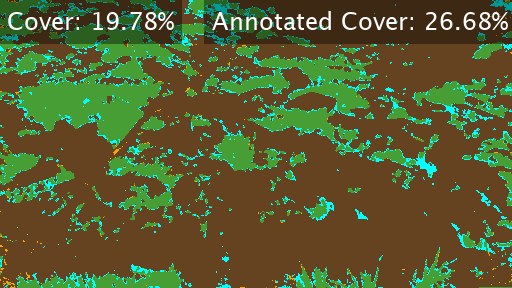} \\
        \vspace{0.1cm}
        \includegraphics[width=.24\textwidth]{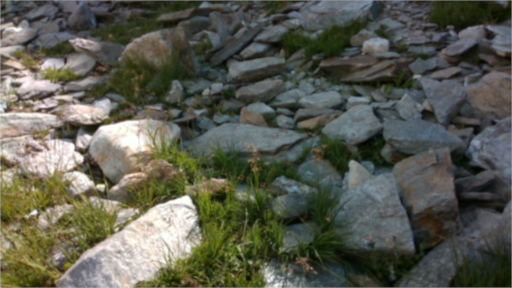}
        \includegraphics[width=.24\textwidth]{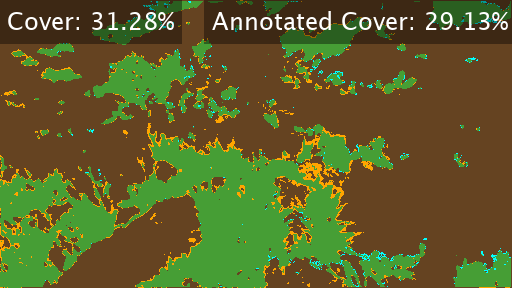} \,
        \includegraphics[width=.24\textwidth]{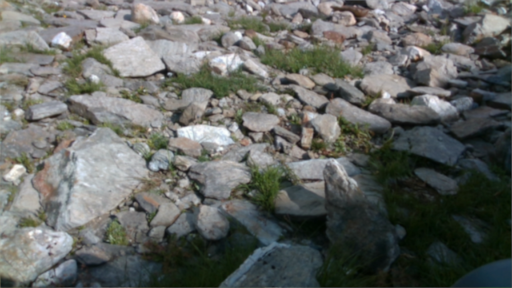}
        \includegraphics[width=.24\textwidth]{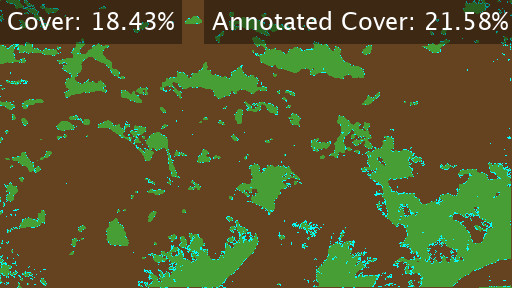}
    \caption{\rev{Comparison of the vegetation cover estimate with manual annotations. Green \textcolor{TP}{\rule{1.5ex}{1.5ex}} does encode vegetation true positive, brown \textcolor{TN}{\rule{1.5ex}{1.5ex}} for background true negative, orange \textcolor{FP}{\rule{1.5ex}{1.5ex}} for false positive, and cyan \textcolor{FN}{\rule{1.5ex}{1.5ex}} for false negative. The estimated vegetation cover with the ExGI method and with the manual annotation are reported in the top left and right corners, respectively.}\label{fig:veg_cov}}
\end{figure*}

In this section, we present the robotic monitoring framework and how it can be instrumental for scree monitoring.
The activity is divided into two phases: (1) the mapping and (2) the autonomous survey.
Thanks to the localization capabilities of the robot, the plot setup and delimitation are no longer necessary.  
Instead, a quick mapping phase is required for the robot's localization during the autonomous operation.
Finally, the autonomous survey is fundamentally different from how it is usually performed by botanists.

\Cref{fig:mission_grid} represents the delimited plot and the starting position of the robot.
The square grid used in the mission setup (in orange) and the waypoint trajectory followed by the robot (in blue) are superimposed on the plot.
\Cref{fig:mission_map} portrays the point cloud obtained from the robot's lidar and the robot's position within the map.
The color of the points encodes the terrain's height.
\Cref{fig:mission_snapshots} shows the real path followed by the robot during the monitoring mission, with several snapshots of the robot superimposed on it.
The mission video is available in the supplementary material, which includes both the mapping and the mission execution (from minute 2:15 to 2:53).

\Cref{fig:depth_images} shows some depth camera images acquired during the mission.
More specifically, \rev{\cref{fig:depth_front_0,fig:depth_front_1}} represent the front camera images, \rev{\cref{fig:depth_rear_0,fig:depth_rear_1}} the rear camera images, \rev{\cref{fig:depth_left_0,fig:depth_left_1}} the left camera images, and \rev{\cref{fig:depth_right_0,fig:depth_right_1}} the right camera images.
\rev{From these photos, it can be observed how the perspective of the images is different from the usual point of view of human botanists, and how the robot height and the camera angles guarantee both a wide field of view and a good visibility of the ground, fundamental for the plant species detection and the vegetation cover estimation.}

\Cref{fig:monitoring_results} reports some metrics acquired during the robotic monitoring.
\Cref{fig:mon_times} \rev{shows} the times of the robotic monitoring and of the traditional one.
These times include the setup phase and the mapping phase for the traditional and the robotic monitoring, respectively. 
It is noteworthy that, during the robotic survey, data is only gathered, since it is stored and analyzed either autonomously or by botanists at any time.
Moreover, the botanists performed a phytosociological survey, which \rev{differs in scope and outputs from the robotic survey}.
\rev{Therefore, these data should not be interpreted as a direct comparison between the two approaches.}
\Cref{fig:mon_inclinations} reports the roll and pitch inclinations of the terrain plane during the monitoring missions.
It can be noted how, at certain points, the total inclination reaches challenging inclinations of about 30°.
\rev{
Additionally, \cref{fig:mon_battery_usage,fig:mon_power_usage,fig:mon_slippage} show the battery usage (both in KW and in percentage), the average power usage, and the normalized slippages during the missions, respectively.
\Cref{fig:mon_battery_usage} shows that the total battery usage during the monitoring missions was reasonably low, allowing for multiple missions on a single battery charge/
Additionally, \cref{fig:mon_power_usage} shows that the average power usage does not vary too much among different missions, regardless of the terrain conditions (i.e., slope~\cref{fig:mon_inclinations} and terrain slippage~\cref{fig:mon_slippage}).
Therefore, although the terrain morphology and traversability heavily affect the robot's locomotion, the control algorithm is able to locomote efficiently without excessive power usage.
}

Finally, in \cref{fig:slip_snap}, a detailed view of the terrain moving under the robot's feet during the monitoring mission is shown.
\rev{The red overlay highlights the terrain movement between two consecutive frames caused by the robot's locomotion.
As can be seen, the terrain moves significantly under the robot's feet.}

\subsection{Vegetation Cover and Plant Species Detection}\label{sec:vegetation_detection}

The vegetation cover estimation is carried out by analyzing the data acquired during the robotic survey.
More specifically, we employed the Excess Green Index (ExGI) (see~\cref{eq:exgi}) for estimating the area occupied by the vegetation.
Some results of the vegetation cover estimate are represented in~\cref{fig:veg_cov}.
Here, we report the results of the ExGI method compared with manual annotations.
The green \textcolor{TP}{\rule{1.5ex}{1.5ex}} represents the vegetation true positive, the brown \textcolor{TN}{\rule{1.5ex}{1.5ex}} for background true negative, the orange \textcolor{FP}{\rule{1.5ex}{1.5ex}} for false positive, and the cyan \textcolor{FN}{\rule{1.5ex}{1.5ex}} for false negative.
In addition, the estimated vegetation cover with the ExGI method and with the manual annotation are reported in the top left and right corners, respectively.
    
\begin{figure*}[htb]
    \centering
    \renewcommand{\arraystretch}{1.3}
    \begin{tabular}[b]{c@{\,}c@{\,}c@{\,}c}
        \includegraphics*[width=0.235\textwidth]{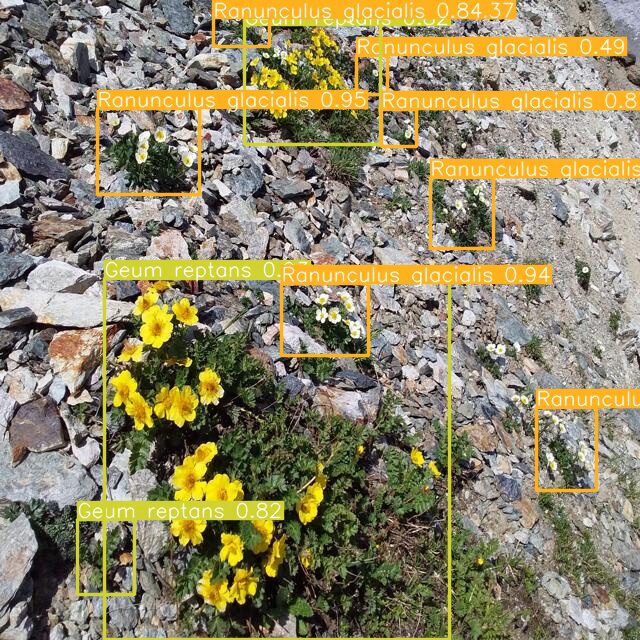} &
        \includegraphics*[width=0.235\textwidth]{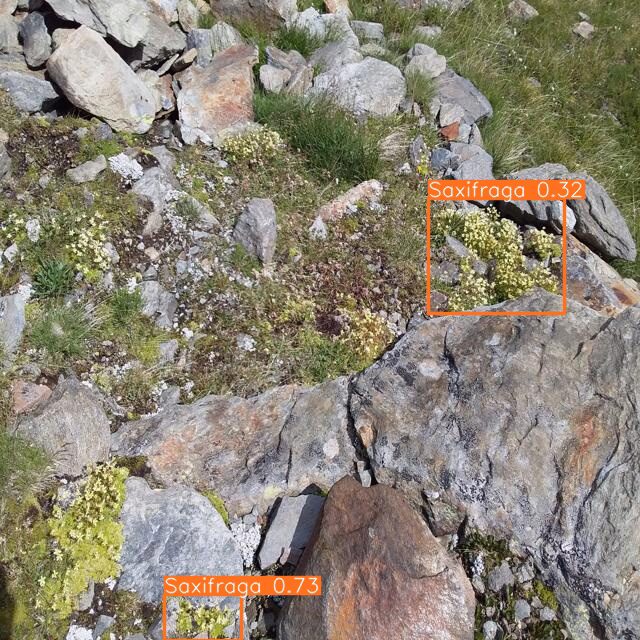} &
        \includegraphics*[width=0.235\textwidth]{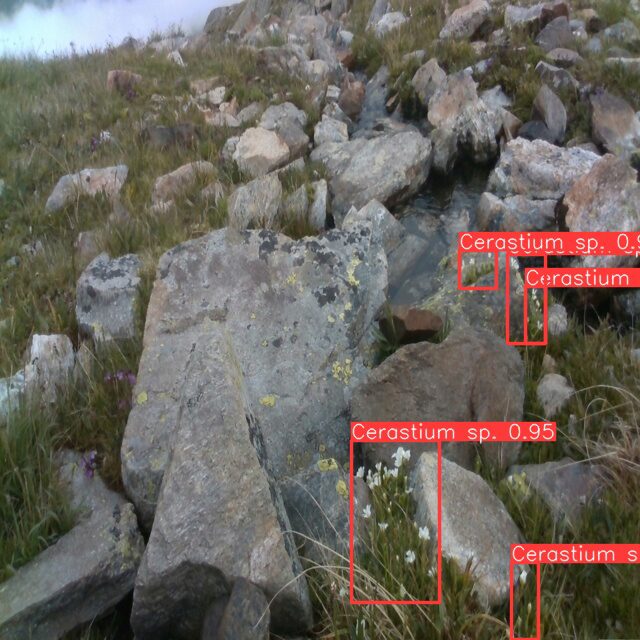} &
        \includegraphics*[width=0.235\textwidth]{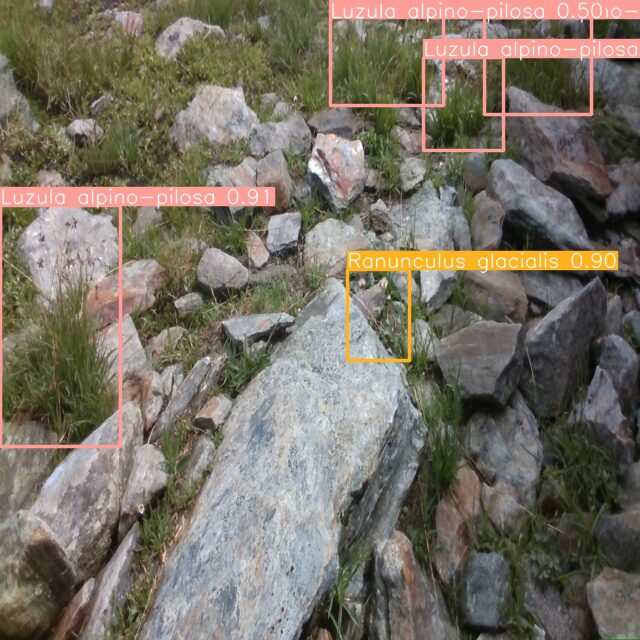}
    \end{tabular}
    \caption{Example images processed by the neural network.}\label{fig:nn_inference}
\end{figure*}

We propose a neural network for performing object detection of the TS and EWS of scree habitats.
More specifically, the selected TS are \emph{Cerastium} spp. (which includes \emph{Cerastium uniflorum} Clairv.\ and \emph{Cerastium pedunculatum} Gaudin), \emph{Geum reptans} L., \emph{Papaver alpinum} L., \emph{Ranunculus glacialis} L., and \emph{Saxifraga bryoides} L.
Conversely, the only recorded EWS is the \emph{Luzula alpinopilosa} Chaix Breistr.
Therefore, the total number of classes is six.

The detection performance of the developed neural network is evaluated using four key performance indicators: precision ($P$), recall ($R$), mean Average Precision (mAP) mAP50, and mAP50-95.
Precision can be thought of as the quality of positive predictions, while recall is the completeness of positive predictions~\cite{van2004geometry}.
Conversely, mAP50 and mAP50-95 also relate to the accuracy of the bounding box compared to the ground-truth box (known as Intersection over Union or Jaccard index)~\cite{padilla2020survey}.

More specifically, the precision and recall are computed as
\begin{equation}
        P = \frac{TP}{TP + FP},
        \quad
        R = \frac{TP}{TP + FN},
\end{equation}
where $TP$ is the number of true positives, $FP$ is the number of false positives, and $FN$ is the number of false negatives.

The Intersection over Union (IoU) is the ratio of the overlap of the predicted ($B_\mathrm{p}$) and ground truth ($B_\mathrm{t}$) boxes over the union of the two boxes, i.e. $\text{IoU} = (B_\mathrm{p} \cap B_\mathrm{t}) / (B_\mathrm{p} \cup B_\mathrm{t})$. 
By choosing a threshold $t \in [0, 1]$, we can compute the precision and the recall for each class by considering a prediction as correct when its IoU is greater than the threshold $t$.
The mAP50 is the Precision with a threshold of 0.5; the mAP50-95 is the average of the precisions for thresholds $t = 0.5, 0.55, \dots, 0.95$, i.e.
\begin{equation}
    \begin{aligned}
        \text{mAP50} &= P(\text{IoU} \geq 0.5), \\
        \text{mAP50-95} &= \frac{1}{10} \sum_{i = 0}^{9} P(\text{IoU} \geq 0.5 + 0.05 i).
    \end{aligned}
\end{equation}

The neural network performance metrics are represented in \cref{tab:nn_performance}.
Some examples of the neural network inference on the data are shown in \cref{fig:nn_inference}.
The output of the neural network consists of the bounding boxes of the detected plants and the corresponding class, which can be displayed on the input images.

\begin{table}[htb]
    \centering
    \caption{Performance of the neural network on the test set.}
    \label{tab:nn_performance}
    \resizebox{0.5\textwidth}{!}{%
    \begin{tabular}[b]{lcccc}
        \toprule
        \textbf{Class} & $\mathbf{Precision}$ & $\mathbf{Recall}$ & \textbf{mAP50} & \textbf{mAP50-95} \\
        \midrule
        All & 0.791 & 0.633 & 0.726 & 0.423 \\
        \emph{Cerastium} & 0.762 & 0.429 & 0.55 & 0.313 \\
        \emph{Geum reptans} & 0.786 & 0.635 & 0.737 & 0.381 \\
        \emph{Papaver alpinum} & 0.74 & 0.712 & 0.743 & 0.363 \\
        \emph{Ranunculus glacialis} L. & 0.758 & 0.738 & 0.809 & 0.489 \\
        \emph{Saxifraga bryoides} L. T. & 0.838 & 0.755 & 0.824 & 0.576 \\
        \emph{Luzula alpinopilosa} & 0.86 & 0.531 & 0.695 & 0.417 \\
        \bottomrule
    \end{tabular}
    }
\end{table}
\section{Discussion}\label{sec:discussion}

In this section, we discuss the results obtained from the field trials.
Afterward, in \cref{sec:limitations}, we highlight the main limitations of the proposed framework and outline future work to address these limitations.

\rev{
The first key takeaway from the field trials performed with ANYmal C is that commercially available legged robots are operationally ready for scree terrains and other very challenging natural environments.
The robot followed challenging paths with several terrain types and characteristics (see~\cref{fig:scree_terrains}) to reach monitoring plots far from each other (as shown in \cref{fig:mission_path_1,fig:mission_path_3}).
During both traveling and autonomous monitoring, the robot successfully navigated steep slopes (up to about $\SI{30}{\degree}$, as shown in the third plot of \cref{fig:mon_inclinations}) and unstable terrains.
}
However, some monitoring plots of interest were avoided for safety reasons when the terrain demanded extreme caution or when unstable screes and block size \rev{were challenging for} the robot's advance.
It is worth noting that ANYmal C used a blind control approach during these field trials.
Conversely, non-blind controllers have recently achieved very promising results on natural environments~\cite{miki2022learning} or in obstacle-rich environments~\cite{hoeller2024anymal}.
Therefore, the integration of such controllers could greatly enhance the robot's locomotion capabilities in challenging terrains.

\rev{
The second key takeaway is that the proposed robotic monitoring framework can effectively collect high-quality plant species data from scree habitats.
The custom neural network for plant species detection achieved promising results, especially when considering the small dataset size with respect to the complexity of the identification task, which, in general, can be challenging and error-prone even for expert botanists~\cite{morrison2016observer}.
As shown in \cref{tab:nn_performance}, the precision of the detection ranged from 86\% for the easier plant species to about 75\% for the more challenging ones.
Similarly, the detected bounding boxes had good quality, as shown by the mAP50 values greater than 0.7 and mAP50-95 greater than 0.35 for most species (in challenging scenarios, mAP50 $\geq 0.6$ and mAP50-95 $\geq 0.35$ are considered good results).
These values are obtained by comparing the inference to the ground truth labels, which were manually annotated by expert botanists.
Note that there are no available neural networks or easily accessible public datasets for scree species detection that could be used for a comparative analysis.
Notably, \cite{stevens2024bioclip} presents a foundation model for plant species classification, which includes mountain vegetation, but it is not suitable for detection.
Therefore, the only way to compare the results is with respect to the manual annotations, as in~\cref{tab:nn_performance}.
}

\rev{
In addition to automated plant detection, the proposed monitoring framework also offers accurate and consistent collection of the vegetation cover and of the terrain stability.
For instance, visual estimation can easily result in very high errors, especially for low coverage values~\cite{morrison2016observer}.
On the other hand, the ExGI method provides a much more reliable estimation of the vegetation cover, as it is based on high-resolution images and does not rely on subjective human judgment or guidelines interpretation.
When compared with manual annotations (\cref{fig:veg_cov}), the ExGI method achieved very good performance, managing to reach percentage errors lower than 1\% in optimal conditions.
In low light conditions, the ExGi and the manual annotations did not match perfectly, having errors of about 3\%.
However, in these cases, manual annotations were also very challenging and unreliable, as the vegetation was not clearly visible.
Similarly, the slippage detection algorithm provides a quantitative measure of the terrain's stability, which is crucial for understanding the habitat's dynamics.
In this case, no quantitative comparison is possible, as traditional monitoring guidelines rely only on a qualitative and subjective assessment of the terrain stability.
Therefore, the proposed slippage metric provides a first step toward a more objective and quantitative assessment of the terrain stability.
}

\rev{
Additionally, one important aspect of the proposed robotic monitoring framework is that it can considerably reduce the on-site workload.
The high execution speed of robotic monitoring (see \cref{fig:mon_times}), coupled with the fact that the mission can be fully automated, means that the robot can be deployed in the field with minimal human supervision and achieve high productivity.
While these results are promising, particular attention must be paid to the specific context and requirements of each monitoring approach.
Given the different nature of the two surveys, both in modalities and in output, it is important to stress the importance of traditional monitoring, while acknowledging the potential benefits of a supplementary robotic approach.
}

\rev{Concluding, compared} with traditional human-only approaches, the proposed methodology reduces on-site workload and enhances data consistency \rev{and quality}.
Nonetheless, trained botanists remain indispensable for comprehensive phytosociological surveys, \rev{while} the robotic system contributes high-frequency, high-resolution observations that can be analyzed offline to strengthen overall habitat assessments.
By integrating these two data \rev{sources and harnessing the knowledge of botanists to aggregate and evaluate the data collected by the robotic platform}, researchers can obtain a more holistic perspective on scree ecosystems\rev{, boosting} monitoring frequency without overburdening field teams.

\subsection{\rev{Main Limitations and Future Work}}
\label{sec:limitations}

\rev{The main limitations of the proposed framework can be divided into three categories: hardware-related, software-related, and integration-related.}

\rev{Hardware-side, the biggest limitation experienced during the field campaign was the overheating of the actuators of the ANYmal C robot.
Temperatures exceeding $\SI{25}{\degree}$, direct sun exposure, and long travels with steep inclinations and demanding terrains challenged the robot's capabilities.
However, the newer version of the robot, ANYmal D, has been equipped with more efficient cooling systems, which will greatly improve the robot's performance in such conditions.
Similarly, many recent legged robots have been designed to operate in more extreme conditions.
Additionally, control strategies that are temperature-aware could further enhance robot capabilities even under more extreme conditions.}

\rev{Software-side, the detection accuracy of the custom neural network was relatively low for some species, particularly the \emph{Cerastium} spp.
This limitation is primarily due to the small size of the training dataset, which does not adequately represent the variability of the species in the field.
Additionally, merging two different species into a single class further complicated the classification.
Finally, the plant morphology and colors of the \emph{Cerastium} spp. further complicates the classification task.}

\rev{
Future work will address several aspects of the proposed framework to enhance its robustness, scalability, and overall performance.
On the robotic side, both the hardware and the software will be improved.
Other than resolving thermal issues, the development of novel bioinspired feet~\cite{catalano2021adaptive} and multimodal locomotion modalities~\cite{ranjan2023design} will be explored.
On the control side, non-blind~\cite{10993458} and terrain-aware~\cite{debenedittis2025soft} controllers also have the potential to enhance locomotion capabilities, particularly in challenging scree terrains.

A more advanced path-planning algorithm will also be developed to optimize obstacle avoidance within the monitoring plot, reduce overall survey time, and improve the quantity and quality of the collected data.
More specifically, plant detection accuracy will be leveraged to dynamically adjust the robot's planned path, ensuring comprehensive coverage of the habitat and accurate detection of all target species.

On the neural network side, efforts will focus on significantly expanding the dataset to improve the model's robustness across all classes, particularly the more challenging ones.
The enhanced dataset will include a broader variety of lighting conditions, environmental variations, and plant morphologies to ensure better generalization.
Furthermore, additional plant species of interest will be incorporated into the dataset, allowing the framework to monitor a wider range of biodiversity.

Great focus will be placed on extracting more global and comprehensive information from the collected data.
Methods to aggregate the plant species detection and avoid double-counting will exploit the spatial distribution of the detected plants.
Additionally, the total vegetation cover of the plot will be estimated using the point cloud data in addition to the camera images, or using image-only data and deep learning methods~\cite{pow3r_cvpr25}.
Finally, new field campaigns will be planned to provide a better understanding of the differences between traditional and robotic monitoring approaches.
}
\section{Conclusion}\label{sec:conclusion}

This study introduced and validated a framework for autonomous robotic monitoring of scree habitats, marking a pioneering application in this challenging field.
Field trials demonstrated that state-of-the-art robots could efficiently navigate steep and unstable terrains, and, in the near future, have the potential to achieve effective coverage of target areas with low battery consumption.
In doing so, it showed the capability to address several longstanding limitations of conventional monitoring methods, notably the time-consuming nature of fieldwork and the difficulty of achieving consistent, repeatable measurements.
Furthermore, it highlighted the main areas of improvement, both from a hardware and a software perspective.

The efficacy of this approach arises from the synergy between three key elements:
\begin{itemize}
    \item \emph{Robotic Data Collection}: Rapid, systematic acquisition of visual and sensor data in sites that are hard to reach or potentially dangerous for human surveyors.  
    \item \emph{Phytosociological Expertise}: Specialist knowledge to confirm plant identities, interpret ecological trends, integrate findings with broader conservation efforts, and fulfill a multifaceted educational role.
    \item \emph{Automated Analysis}: Neural-network-based detection of both TS and EWS, paired with quantitative indicators such as vegetation cover and a slippage metric.
\end{itemize}

Taken together, these advances enable a more frequent, cost-effective, and objective mode of habitat assessment in high-altitude ecosystems.  
Looking ahead, the continued evolution of robotic hardware and AI-based processing promises even broader adoption of automated field methods, particularly in remote or risky environments.
By leveraging the strengths of both robotics and expert ecological insight, future investigations can expand the scope and detail of habitat monitoring, ultimately enriching our understanding of biodiversity patterns and environmental change in mountainous regions.

\printbibliography

@article{nitzu2014scree,
  title={Scree habitats: ecological function, species conservation and spatial-temporal variation in the arthropod community},
  author={Nitzu, Eugen and Nae, A and B{\u{a}}ncil{\u{a}}, R and Popa, I and Giurginca, A and Pl{\u{a}}ia{\c{s}}u, R},
  journal={Systematics and Biodiversity},
  volume={12},
  number={1},
  pages={65--75},
  year={2014},
  publisher={Taylor \& Francis}
}

@article{fragniere2020climate,
  title={Climate change and alpine screes: no future for glacial relict Papaver occidentale (Papaveraceae) in Western Prealps},
  author={Fragni{\`e}re, Yann and Pittet, Lo{\"\i}c and Cl{\'e}ment, Beno{\^\i}t and B{\'e}trisey, S{\'e}bastien and Gerber, Emanuel and Ronikier, Micha{\l} and Parisod, Christian and Kozlowski, Gregor},
  journal={Diversity},
  volume={12},
  number={9},
  pages={346},
  year={2020},
  publisher={MDPI}
}

@article{bhardwaj2016uavs,
	title={UAVs as remote sensing platform in glaciology: Present applications and future prospects},
	author={Bhardwaj, Anshuman and Sam, Lydia and Mart{\'\i}n-Torres, F Javier and Kumar, Rajesh and others},
	journal={Remote sensing of environment},
	volume={175},
	pages={196--204},
	year={2016},
	publisher={Elsevier}
}

@article{van2018dawning,
	title={The dawning of the ethics of environmental robots},
	author={Van Wynsberghe, Aimee and Donhauser, Justin},
	journal={Science and engineering ethics},
	volume={24},
	number={6},
	pages={1777--1800},
	year={2018},
	publisher={Springer}
}

@article{directive1992council,
	title={{Council Directive 92/43/ EEC of 21 May 1992 on the conservation of natural habitats and of wild fauna and flora.}},
	author={{European Commission}},
	journal={Official Journal of the European Union},
	volume={206},
	pages={7--50},
	year={1992}
}

@ARTICLE{angelini2023robotic,
  author={Angelini, Franco and Angelini, Pierangela and Angiolini, Claudia and Bagella, Simonetta and Bonomo, Fabio and Caccianiga, Marco and Santina, Cosimo Della and Gigante, Daniela and Hutter, Marco and Nanayakkara, Thrishantha and Remagnino, Paolo and Torricelli, Diego and Garabini, Manolo},
  journal={IEEE Access}, 
  title={Robotic Monitoring of Habitats: The Natural Intelligence Approach}, 
  year={2023},
  volume={11},
  number={},
  pages={72575-72591},
  doi={10.1109/ACCESS.2023.3294276}}

@article{kati2015challenge,
  title={The challenge of implementing the European network of protected areas Natura 2000},
  author={Kati, Vassiliki and Hovardas, Tasos and Dieterich, Martin and Ibisch, Pierre L and Mihok, Barbara and Selva, Nuria},
  journal={Conservation Biology},
  volume={29},
  number={1},
  pages={260--270},
  year={2015},
  publisher={Wiley Online Library}
}

@article{bonari2021shedding,
  title={Shedding light on typical species: implications for habitat monitoring},
  author={Bonari, Gianmaria and Fantinato, Edy and Lazzaro, Lorenzo and Sperandii, Marta Gaia and Acosta, Alicia Teresa Rosario and Allegrezza, Marina and Assini, Silvia and Caccianiga, Marco and Di Cecco, Valter and Frattaroli, Annarita and others},
  journal={Plant Sociology},
  volume={58},
  number={1},
  pages={157--166},
  year={2021},
  publisher={Societa Italiana di Scienza della Vegetazione}
}

@article{gigante2016methodological,
  title={A methodological protocol for Annex I Habitats monitoring: the contribution of Vegetation science.},
  author={Gigante, Daniela and Attorre, F and Venanzoni, Roberto and Acosta, ATR and Agrillo, E and Aleffi, Michele and Alessi, N and Allegrezza, Marina and Angelini, P and Angiolini, C and others},
  journal={Plant Sociology},
  volume={53},
  number={2},
  pages={77--87},
  year={2016}
}

@article{dunbabin2012robots,
  title={Robots for environmental monitoring: Significant advancements and applications},
  author={Dunbabin, Matthew and Marques, Lino},
  journal={IEEE Robotics \& Automation Magazine},
  volume={19},
  number={1},
  pages={24--39},
  year={2012},
  publisher={IEEE}
}

@article{puliti2015inventory,
  title={Inventory of small forest areas using an unmanned aerial system},
  author={Puliti, Stefano and {\O}rka, Hans Ole and Gobakken, Terje and N{\ae}sset, Erik},
  journal={Remote Sensing},
  volume={7},
  number={8},
  pages={9632--9654},
  year={2015},
  publisher={MDPI}
}

@article{michez2016classification,
  title={Classification of riparian forest species and health condition using multi-temporal and hyperspatial imagery from unmanned aerial system},
  author={Michez, Adrien and Pi{\'e}gay, Herv{\'e} and Lisein, Jonathan and Claessens, Hugues and Lejeune, Philippe},
  journal={Environmental monitoring and assessment},
  volume={188},
  pages={1--19},
  year={2016},
  publisher={Springer}
}

@article{koh2012dawn,
  title={Dawn of drone ecology: low-cost autonomous aerial vehicles for conservation},
  author={Koh, Lian Pin and Wich, Serge A},
  journal={Tropical conservation science},
  volume={5},
  number={2},
  pages={121--132},
  year={2012},
  publisher={SAGE Publications Sage CA: Los Angeles, CA}
}

@article{paneque2014small,
  title={Small drones for community-based forest monitoring: An assessment of their feasibility and potential in tropical areas},
  author={Paneque-G{\'a}lvez, Jaime and McCall, Michael K and Napoletano, Brian M and Wich, Serge A and Koh, Lian Pin},
  journal={Forests},
  volume={5},
  number={6},
  pages={1481--1507},
  year={2014},
  publisher={Multidisciplinary Digital Publishing Institute}
}

@article{sankey2017uav,
  title={UAV lidar and hyperspectral fusion for forest monitoring in the southwestern USA},
  author={Sankey, Temuulen and Donager, Jonathon and McVay, Jason and Sankey, Joel B},
  journal={Remote Sensing of Environment},
  volume={195},
  pages={30--43},
  year={2017},
  publisher={Elsevier}
}

@article{seier2017uas,
  title={UAS-based change detection of the glacial and proglacial transition zone at Pasterze Glacier, Austria},
  author={Seier, Gernot and Kellerer-Pirklbauer, Andreas and Wecht, Matthias and Hirschmann, Simon and Kaufmann, Viktor and Lieb, Gerhard K and Sulzer, Wolfgang},
  journal={Remote Sensing},
  volume={9},
  number={6},
  pages={549},
  year={2017},
  publisher={MDPI}
}

@article{mayr2018disturbance,
  title={Disturbance feedbacks on the height of woody vegetation in a savannah: a multi-plot assessment using an unmanned aerial vehicle (UAV)},
  author={Mayr, Manuel J and Mal{\ss}, Sophia and Ofner, Elisabeth and Samimi, Cyrus},
  journal={International Journal of Remote Sensing},
  volume={39},
  number={14},
  pages={4761--4785},
  year={2018},
  publisher={Taylor \& Francis}
}

@article{woodget2017drones,
  title={Drones and digital photogrammetry: from classifications to continuums for monitoring river habitat and hydromorphology},
  author={Woodget, Amy S and Austrums, Robbie and Maddock, Ian P and Habit, Evelyn},
  journal={Wiley Interdisciplinary Reviews: Water},
  volume={4},
  number={4},
  pages={e1222},
  year={2017},
  publisher={Wiley Online Library}
}

@inproceedings{bapna1998atacama,
  title={The atacama desert trek: Outcomes},
  author={Bapna, Deepak and Rollins, Eric and Murphy, John and Maimone, E and Whittaker, William and Wettergreen, David},
  booktitle={Proceedings. 1998 IEEE International Conference on Robotics and Automation (Cat. No. 98CH36146)},
  volume={1},
  pages={597--604},
  year={1998},
  organization={IEEE}
}

@article{muscato2003robovolc,
  title={ROBOVOLC: A robot for volcano exploration result of first test campaign},
  author={Muscato, Giovanni and Caltabiano, Daniele and Guccione, Salvatore and Longo, Domenico and Coltelli, Mauro and Cristaldi, A and Pecora, Emilio and Sacco, Vincenzo and Sim, Patrick and Virk, Gurvinder S and others},
  journal={Industrial Robot: An International Journal},
  volume={30},
  number={3},
  pages={231--242},
  year={2003},
  publisher={MCB UP Ltd}
}

@article{notomista2019slothbot,
  title={The SlothBot: A novel design for a wire-traversing robot},
  author={Notomista, Gennaro and Emam, Yousef and Egerstedt, Magnus},
  journal={IEEE Robotics and Automation Letters},
  volume={4},
  number={2},
  pages={1993--1998},
  year={2019},
  publisher={IEEE}
}

@article{bares1999dante,
  title={Dante II: Technical description, results, and lessons learned},
  author={Bares, John E and Wettergreen, David S},
  journal={The International Journal of robotics research},
  volume={18},
  number={7},
  pages={621--649},
  year={1999},
  publisher={SAGE Publications}
}

@article{semini2011design,
  title={Design of HyQ--a hydraulically and electrically actuated quadruped robot},
  author={Semini, Claudio and Tsagarakis, Nikos G and Guglielmino, Emanuele and Focchi, Michele and Cannella, Ferdinando and Caldwell, Darwin G},
  journal={Proceedings of the Institution of Mechanical Engineers, Part I: Journal of Systems and Control Engineering},
  volume={225},
  number={6},
  pages={831--849},
  year={2011},
  publisher={SAGE Publications Sage UK: London, England}
}

@inproceedings{hutter2016anymal,
  title={Anymal-a highly mobile and dynamic quadrupedal robot},
  author={Hutter, Marco and Gehring, Christian and Jud, Dominic and Lauber, Andreas and Bellicoso, C Dario and Tsounis, Vassilios and Hwangbo, Jemin and Bodie, Karen and Fankhauser, Peter and Bloesch, Michael and others},
  booktitle={2016 IEEE/RSJ international conference on intelligent robots and systems (IROS)},
  pages={38--44},
  year={2016},
  organization={IEEE}
}

@article{angelini2023alpine,
  title={Robotic monitoring of Alpine screes: a dataset from the EU Natura2000 habitat 8110 in the Italian Alps},
  author={Angelini, Franco and Pollayil, Mathew J and Valle, Barbara and Borgatti, Marina Serena and Caccianiga, Marco and Garabini, Manolo},
  journal={Scientific Data},
  volume={10},
  number={1},
  pages={855},
  year={2023},
  publisher={Nature Publishing Group UK London}
}

@book{elzinga1998measuring,
  title={Measuring \& monitoring plant populations},
  author={Elzinga, Caryl L and Salzer, Daniel W},
  year={1998},
  publisher={US Department of the Interior, Bureau of Land Management}
}

@article{ellwanger2018current,
  title={Current status of habitat monitoring in the European Union according to Article 17 of the Habitats Directive, with an emphasis on habitat structure and functions and on Germany},
  author={Ellwanger, G{\"o}tz and Runge, Stephan and Wagner, Melanie and Ackermann, Werner and Neukirchen, Melanie and Frederking, Wenke and M{\"u}ller, Christina and Ssymank, Axel and Sukopp, Ulrich},
  journal={Nature Conservation},
  volume={29},
  pages={57--78},
  year={2018},
  publisher={Pensoft Publishers}
}

@software{yolo,
author = {Jocher, Glenn and Chaurasia, Ayush and Qiu, Jing},
license = {AGPL-3.0},
month = jan,
title = {{Ultralytics YOLO}},
url = {https://github.com/ultralytics/ultralytics},
version = {8.0.0},
year = {2023}
}

@article{larrinaga2019greenness,
  title={Greenness indices from a low-cost UAV imagery as tools for monitoring post-fire forest recovery},
  author={Larrinaga, Asier R and Brotons, Lluis},
  journal={Drones},
  volume={3},
  number={1},
  pages={6},
  year={2019},
  publisher={MDPI}
}

@article{miki2022learning,
  title={Learning robust perceptive locomotion for quadrupedal robots in the wild},
  author={Miki, Takahiro and Lee, Joonho and Hwangbo, Jemin and Wellhausen, Lorenz and Koltun, Vladlen and Hutter, Marco},
  journal={Science Robotics},
  volume={7},
  number={62},
  pages={eabk2822},
  year={2022},
  publisher={American Association for the Advancement of Science}
}

@misc{hab_dir_art_17,
  title = {Habitat Directive Art. 17},
  note = {\url{https://nature-art17.eionet.europa.eu/article17/habitat/report/}}
}

@article{lee2020learning,
  title={Learning quadrupedal locomotion over challenging terrain},
  author={Lee, Joonho and Hwangbo, Jemin and Wellhausen, Lorenz and Koltun, Vladlen and Hutter, Marco},
  journal={Science robotics},
  volume={5},
  number={47},
  pages={eabc5986},
  year={2020},
  publisher={American Association for the Advancement of Science}
}

@book{van2004geometry,
  title={The geometry of information retrieval},
  author={Van Rijsbergen, Cornelis Joost},
  year={2004},
  publisher={Cambridge University Press}
}

@inproceedings{padilla2020survey,
  title={A survey on performance metrics for object-detection algorithms},
  author={Padilla, Rafael and Netto, Sergio L and Da Silva, Eduardo AB},
  booktitle={2020 international conference on systems, signals and image processing (IWSSIP)},
  pages={237--242},
  year={2020},
  organization={IEEE}
}

@article{catalano2021adaptive,
  title={Adaptive feet for quadrupedal walkers},
  author={Catalano, Manuel Giuseppe and Pollayil, Mathew Jose and Grioli, Giorgio and Valsecchi, Giorgio and Kolvenbach, Hendrik and Hutter, Marco and Bicchi, Antonio and Garabini, Manolo},
  journal={IEEE Transactions on Robotics},
  volume={38},
  number={1},
  pages={302--316},
  year={2021},
  publisher={IEEE}
}

@article{camurri2020pronto,
  title={Pronto: A multi-sensor state estimator for legged robots in real-world scenarios},
  author={Camurri, Marco and Ramezani, Milad and Nobili, Simona and Fallon, Maurice},
  journal={Frontiers in Robotics and AI},
  volume={7},
  pages={68},
  year={2020},
  publisher={Frontiers Media SA}
}

@article{fernandez2023exploring,
  title={Exploring the trade-off between performance and annotation complexity in semantic segmentation},
  author={Fern{\'a}ndez-Moreno, Marta and Lei, Bo and Holm, Elizabeth A and Mesejo, Pablo and Moreno, Ra{\'u}l},
  journal={Engineering Applications of Artificial Intelligence},
  volume={123},
  pages={106299},
  year={2023},
  publisher={Elsevier}
}

@article{lathrop2014comparison,
  title={Comparison of remotely-sensed surveys vs. in situ plot-based assessments of sea grass condition in Barnegat Bay-Little Egg Harbor, New Jersey USA},
  author={Lathrop, Richard G and Haag, Scott M and Merchant, Daniel and Kennish, Michael J and Fertig, Benjamin},
  journal={Journal of Coastal Conservation},
  volume={18},
  pages={299--308},
  year={2014},
  publisher={Springer}
}

@ARTICLE{ranjan2023design,
  author={Ranjan, Alok and Angelini, Franco and Nanayakkara, Thrishantha and Garabini, Manolo},
  journal={IEEE/ASME Transactions on Mechatronics}, 
  title={Design Guidelines for Bioinspired Adaptive Foot for Stable Interaction With the Environment}, 
  year={2024},
  volume={29},
  number={2},
  pages={843-855},
  doi={10.1109/TMECH.2023.3326602}}

@ARTICLE{debenedittis2025soft,
  author={De Benedittis, Davide and Angelini, Franco and Garabini, Manolo},
  journal={IEEE Transactions on Systems, Man, and Cybernetics: Systems}, 
  title={Soft Bilinear Inverted Pendulum: A Model to Enable Locomotion With Soft Contacts}, 
  year={2025},
  volume={55},
  number={2},
  pages={1478-1491},
  doi={10.1109/TSMC.2024.3504342}}

@article{choi2023learning,
  title={Learning quadrupedal locomotion on deformable terrain},
  author={Choi, Suyoung and Ji, Gwanghyeon and Park, Jeongsoo and Kim, Hyeongjun and Mun, Juhyeok and Lee, Jeong Hyun and Hwangbo, Jemin},
  journal={Science Robotics},
  volume={8},
  number={74},
  pages={eade2256},
  year={2023},
  publisher={American Association for the Advancement of Science}
}

@INPROCEEDINGS{zhang2023optimal,
  author={Zhang, Yifan and Liu, Hui and Han, Lijin},
  booktitle={2023 5th International Conference on Robotics, Intelligent Control and Artificial Intelligence (RICAI)}, 
  title={Optimal Global Path Planning for Wheel-Legged Robot on Mountain Terrain}, 
  year={2023},
  volume={},
  number={},
  pages={153-157},
  doi={10.1109/RICAI60863.2023.10488967}}

@article{change2007climate,
  title={Climate change 2007: the physical science basis},
  author={Change, Intergovernmental Panel On Climate},
  journal={Agenda},
  volume={6},
  number={07},
  pages={333},
  year={2007}
}

@article{watson2019summary,
  title={Summary for policymakers of the global assessment report on biodiversity and ecosystem services of the Intergovernmental Science-Policy Platform on Biodiversity and Ecosystem Services},
  author={Watson, Robert and Baste, Ivar and Larigauderie, Anne and Leadley, Paul and Pascual, Unai and Baptiste, Brigitte and Demissew, Sebsebe and Dziba, Luthando and Erpul, G{\"u}nay and Fazel, Asghar and others},
  journal={IPBES Secretariat: Bonn, Germany},
  pages={22--47},
  year={2019}
}

@article{biondi2010manuale,
  title={Manuale Italiano di interpretazione degli habitat (Direttiva 92/43 CEE)},
  author={Biondi, E and Blasi, C and Burrascano, S and Casavecchia, S and Copiz, R and Del Vico, E and Galdenzi, D and Gigante, D and Lasen, C and Spampinato, G and others},
  year={2010}
}

@article{hoeller2024anymal,
  title={Anymal parkour: Learning agile navigation for quadrupedal robots},
  author={Hoeller, David and Rudin, Nikita and Sako, Dhionis and Hutter, Marco},
  journal={Science Robotics},
  volume={9},
  number={88},
  pages={eadi7566},
  year={2024},
  publisher={American Association for the Advancement of Science}
}

@article{dilorenzo2025robotic,
  title={Robotic monitoring of European habitats: a labeled dataset for plant detection in Annex I habitats of Italy},
  author={Di Lorenzo, Giovanni and Angelini, Franco and Pierallini, Michele and Tolomei, Simone and De Benedittis, Davide and Denaro, Agnese and Rivieccio, Giovanni and Caria, Maria Carmela and Bonini, Federica and Grassi, Anna and others},
  journal={Scientific Data},
  volume={12},
  number={1},
  pages={1--24},
  year={2025},
  publisher={Nature Publishing Group}
}

@misc{ipbes2019global,
  title={Global assessment report on biodiversity and ecosystem services of the Intergovernmental Science-Policy Platform on Biodiversity and Ecosystem Services},
  author={IPBES, BrondizioES and others},
  journal={IPBES secretariat},
  pages={1148},
  year={2019},
  publisher={IPBES Bonn, Germany}
}

@article{afridi2025impact,
  title={Impact of drone disturbances on wildlife: A review},
  author={Afridi, Saadia and Laporte-Devylder, Lucie and Maalouf, Guy and Kline, Jenna M and Penny, Samuel G and Hlebowicz, Kasper and Cawthorne, Dylan and Lundquist, Ulrik Pagh Schultz},
  journal={Drones},
  volume={9},
  number={4},
  pages={311},
  year={2025},
  publisher={MDPI}
}

@article{ding2023robust,
  title={Robust jumping with an articulated soft quadruped via trajectory optimization and iterative learning},
  author={Ding, Jiatao and van L{\"o}ben Sels, Mees A and Angelini, Franco and Kober, Jens and Della Santina, Cosimo},
  journal={IEEE Robotics and Automation Letters},
  volume={9},
  number={1},
  pages={255--262},
  year={2023},
  publisher={IEEE}
}

@inproceedings{cheng2024quadruped,
  title={Quadruped robot traversing 3D complex environments with limited perception},
  author={Cheng, Yi and Liu, Hang and Pan, Guoping and Liu, Houde and Ye, Linqi},
  booktitle={2024 IEEE/RSJ International Conference on Intelligent Robots and Systems (IROS)},
  pages={9074--9081},
  year={2024},
  organization={IEEE}
}

@ARTICLE{chhatoi2023optimal,
  author={Chhatoi, Saroj Prasad and Pierallini, Michele and Angelini, Franco and Mastalli, Carlos and Garabini, Manolo},
  journal={IEEE Transactions on Robotics}, 
  title={Optimal Control for Articulated Soft Robots}, 
  year={2023},
  volume={39},
  number={5},
  pages={3671-3685},
  doi={10.1109/TRO.2023.3288837}}

@inproceedings{pow3r_cvpr25,
  title={Pow3R: Empowering Unconstrained 3D  Reconstruction with Camera and Scene Priors}, 
  author={Wonbong Jang and Philippe Weinzaepfel and Vincent Leroy and Lourdes Agapito and Jerome Revaud},
  booktitle = {CVPR},
  year = {2025}
}

@inproceedings{stevens2024bioclip,
  title={Bioclip: A vision foundation model for the tree of life},
  author={Stevens, Samuel and Wu, Jiaman and Thompson, Matthew J and Campolongo, Elizabeth G and Song, Chan Hee and Carlyn, David Edward and Dong, Li and Dahdul, Wasila M and Stewart, Charles and Berger-Wolf, Tanya and others},
  booktitle={Proceedings of the IEEE/CVF conference on computer vision and pattern recognition},
  pages={19412--19424},
  year={2024}
}

@article{morrison2016observer,
  title={Observer error in vegetation surveys: a review},
  author={Morrison, Lloyd W},
  journal={Journal of Plant Ecology},
  volume={9},
  number={4},
  pages={367--379},
  year={2016},
  publisher={Oxford University Press UK}
}

@article{valle2022biodiversity,
  title={Biodiversity and ecology of plants and arthropods on the last preserved glacier of the Apennines mountain chain (Italy)},
  author={Valle, Barbara and di Musciano, Michele and Gobbi, Mauro and Bonelli, Marco and Colonnelli, Enzo and Gardini, Giulio and Migliorini, Massimo and Pantini, Paolo and Zanetti, Adriano and Berrilli, Emanuele and others},
  journal={The Holocene},
  volume={32},
  number={8},
  pages={853--865},
  year={2022},
  publisher={SAGE Publications Sage UK: London, England}
}

@article{dengler2016phytosociology,
  title={Phytosociology.--International Encyclopedia of Geography: People, the Earth},
  author={Dengler, J},
  journal={Environment and Technology},
  volume={12},
  pages={1--6},
  year={2016}
}

@ARTICLE{10993458,
  author={Tolomei, Simone and Belter, Dominik and Bednarek, Jakub and Angelini, Franco and Garabini, Manolo},
  journal={IEEE Transactions on Systems, Man, and Cybernetics: Systems}, 
  title={Learning-Based Foot-Shape-Aware Foothold Selection for Quadrupedal Robots}, 
  year={2025},
  volume={},
  number={},
  pages={1-14},
  doi={10.1109/TSMC.2025.3563974}}

@article{pringle2025opportunities,
  title={Opportunities and challenges for monitoring terrestrial biodiversity in the robotics age},
  author={Pringle, Stephen and Dallimer, Martin and Goddard, Mark A and Le Goff, L{\'e}ni K and Hart, Emma and Langdale, Simon J and Fisher, Jessica C and Abad, Sara-Adela and Ancrenaz, Marc and Angeoletto, Fabio and others},
  journal={Nature ecology \& evolution},
  pages={1--12},
  year={2025},
  publisher={Nature Publishing Group UK London}
}

@inproceedings{margolis2023walk,
  title={Walk these ways: Tuning robot control for generalization with multiplicity of behavior},
  author={Margolis, Gabriel B and Agrawal, Pulkit},
  booktitle={Conference on Robot Learning},
  pages={22--31},
  year={2023},
  organization={PMLR}
}

\vfill\pagebreak

\end{document}